%% file: main.tex
\documentclass[a4paper,fleqn]{cas-dc}

\usepackage[numbers]{natbib}
\usepackage{graphicx}
\usepackage{amsmath,amssymb}
\usepackage{xcolor}
\usepackage{tikz}
\usetikzlibrary{arrows.meta,decorations.pathreplacing,positioning}
\usepackage{comment}
\usepackage{microtype}
\usepackage{placeins}

\begin{document}
\let\WriteBookmarks\relax
\def\floatpagepagefraction{1}
\def\textpagefraction{.001}

\shorttitle{A Probabilistic Perspective on  Wasserstein-Based Evidential Uncertainty for OOD Segmentation}
\shortauthors{Brosch et al.}

\title[mode=title]{A Probabilistic Perspective on  Wasserstein-Based Evidential Uncertainty for Out-of-Distribution Segmentation}

\author[1]{Arnold Brosch}
\ead{arnold.brosch@hhu.de}
\credit{Conceptualization, Methodology, Software, Validation,
Formal analysis, Investigation, Data curation,
Writing -- original draft, Writing -- review \& editing,
Visualization}
\cormark[1]

\author[2]{Abdelrahman Eldesokey}[orcid=0000-0003-3292-7153]
\ead{aeldesokey@hbku.edu.qa}
\credit{Conceptualization, Software}
\author[3]{Michael Felsberg}[orcid=0000-0002-6096-3648]
\ead{michael.felsberg@liu.se}
\credit{Conceptualization, Writing – review \& editing}
\author[1]{Kira Maag}[orcid=0000-0003-1767-0476]
\ead{kira.maag@hhu.de}
\credit{Conceptualization, Writing – review \& editing, Supervision, Resources}


\affiliation[1]{
    organization={Heinrich-Heine-University D\"usseldorf},
    addressline={Universit\"atsstraße 1},
    postcode={40225},
    city={D\"usseldorf},
    country={Germany}
}

\affiliation[2]{
  organization={Hamad Bin Khalifa University (HBKU)},
  addressline={Education City, Gate 8, P.O. Box 5825},
  city={Doha},
  country={Qatar}
}

\affiliation[3]{
  organization={Link\"oping University},
  postcode={SE-581 83},
  city={Link\"oping},
  country={Sweden}
}

\cortext[1]{Corresponding author}

\begin{abstract}
Semantic segmentation networks operate on a fixed set of classes and therefore fail when out-of-distribution (OOD) objects appear during deployment, a critical limitation for safety-critical applications such as autonomous driving. Reliably identifying OOD objects requires well-calibrated epistemic uncertainty, yet common softmax-based confidence scores remain overconfident, while Bayesian alternatives such as Monte Carlo dropout or deep ensembles require costly repeated forward passes. Evidential Deep Learning (EDL) offers an efficient alternative by modeling class probabilities as a Dirichlet distribution learned from a single deterministic forward pass. Existing EDL formulations rely on Euclidean objectives that push predictions towards the simplex vertices, encouraging overconfidence rather than preserving uncertainty for unfamiliar inputs. We instead employ Wasserstein-based objectives, which respect the geometry of the probability simplex, and study the influence of the Wasserstein order on segmentation accuracy and OOD detection within a unified evidential framework. We evaluate this framework on a convolutional (DeepLabV3+) and a transformer-based (SegFormer) architecture on the SegmentMeIfYouCan benchmark, including LostAndFound, RoadObstacle21, RoadAnomaly21, and Fishyscapes. Our results show the optimal Wasserstein order is architecture-dependent: second-order objectives dominate on the convolutional backbone, third-order objectives on the transformer backbone, and our framework surpasses comparable baselines on most metrics, with a single deterministic forward pass.
\end{abstract}

\begin{highlights}
\item Wasserstein objectives provide geometry-aware evidential OOD segmentation.
\item The preferred Wasserstein order depends on the network architecture.
\item Requires no auxiliary models, gradients, or additional OOD training data.
\end{highlights}


\maketitle

\section{Introduction}

Deep neural networks (DNNs) have become the dominant paradigm for a wide range of computer vision tasks, including image classification \citep{he2016residual}, object detection \cite{redmon2015unified}, and semantic segmentation \cite{Chen2018_deeplab}.  Semantic segmentation aims to assign a semantic class label to every pixel of an image, thereby enabling dense scene understanding. In safety-critical domains such as autonomous driving, semantic segmentation constitutes a fundamental component of the perception pipeline, providing detailed information about the surrounding environment and supporting downstream planning and decision-making processes. Well performing architectures, including DeepLabV3+ \citep{Chen2018_deeplab} and SegFormer \citep{xie2021segformersimpleefficientdesign}, 
have demonstrated strong performance on large-scale benchmarks and have significantly advanced the state-of-the-art in semantic scene understanding.

\begin{figure}
\centering
\includegraphics[width=0.7\linewidth]{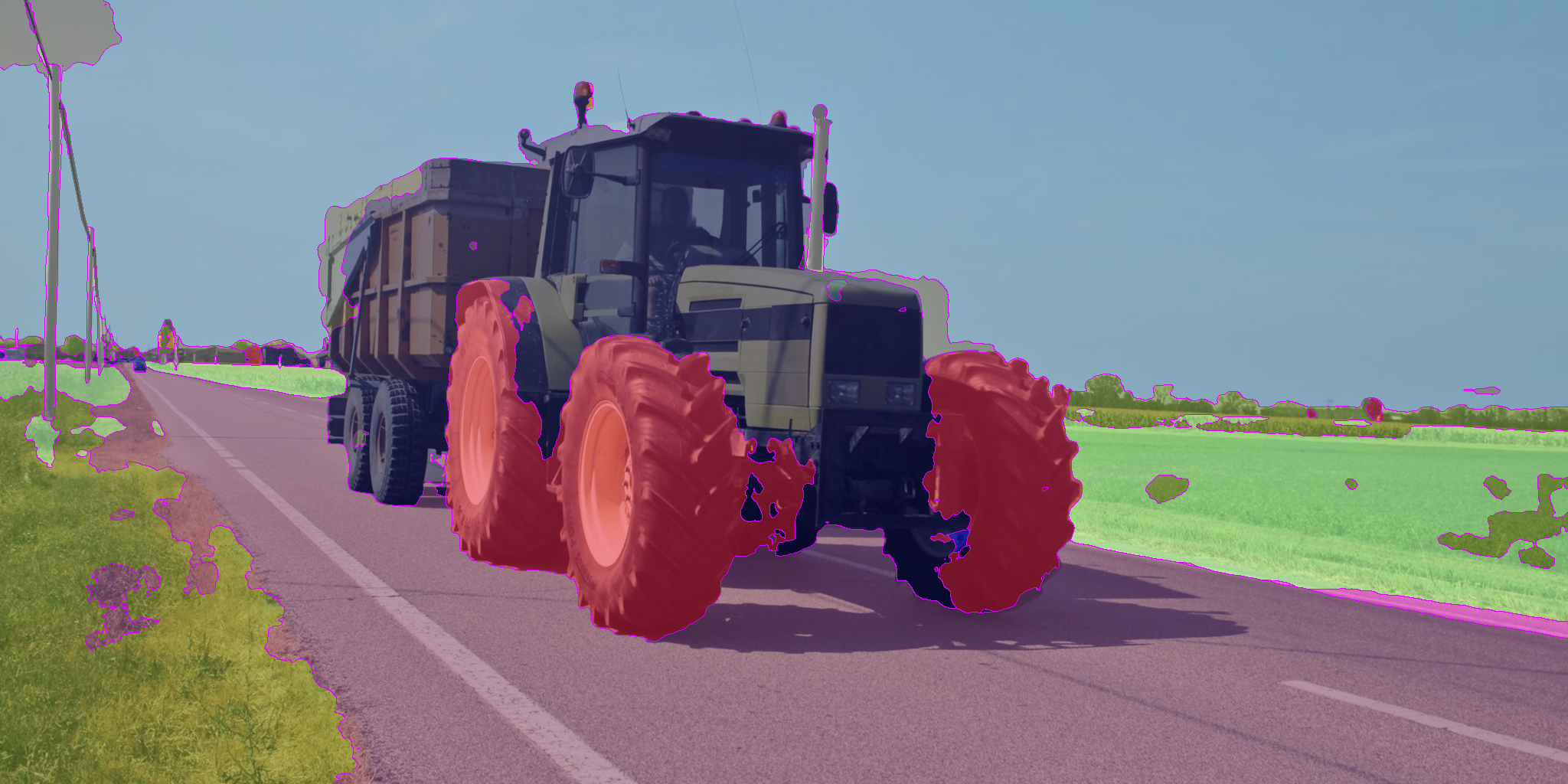} \\
\includegraphics[width=0.7\linewidth]{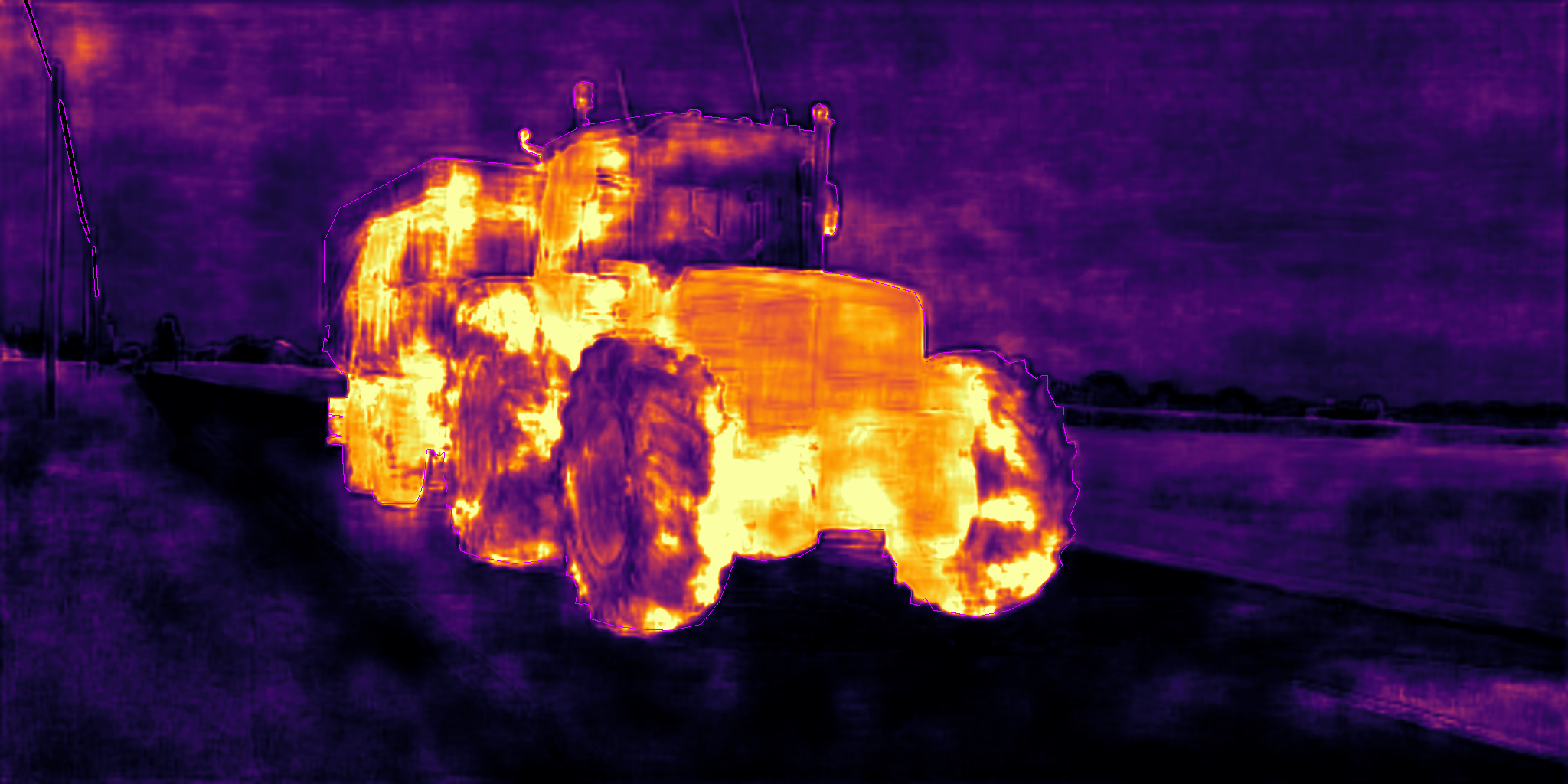}
\caption{Top: Semantic segmentation by a deep neural network. Bottom: Evidential uncertainty heatmap obtained by our method. Brighter colors correspond to higher uncertainty. 
}
\label{fig:example_intro}
\end{figure}

Despite these advances, DNNs remain fundamentally constrained by the data used during training. Semantic segmentation models are typically trained on large annotated datasets that represent only a limited subset of the visual variability encountered in real-world environments. As a result, objects, structures, or scene configurations that are absent from the training distribution may occur during deployment. Such observations are commonly referred to as out-of-distribution (OOD) samples \cite{Chan2021_smiyc,10.1007/978-3-031-26348-4_28}. When confronted with OOD objects, segmentation networks frequently exhibit a substantial degradation in performance and may assign highly confident but incorrect predictions to previously unseen content (see Fig.~\ref{fig:example_intro}). 
In safety-critical applications, such behavior can have severe consequences, as unknown objects may be misclassified as semantically meaningful classes and consequently result in dangerous downstream decisions.

The task of OOD segmentation therefore extends beyond accurate segmentation of known categories and additionally requires reliable identification of regions that do not belong to the learned training distribution.
A central challenge in this context is the estimation of epistemic uncertainty, which arises from insufficient knowledge of the model rather than from noise inherent to the input data (i.e., aleatoric uncertainty).
Ideally, segmentation models should express increased uncertainty whenever unfamiliar or ambiguous observations are encountered. However, uncertainty measures derived from softmax outputs, such as maximum softmax probability \citep{Hendrycks2016} or predictive entropy, are known to be poorly calibrated and often remain overconfident for OOD samples \citep{Hein}. Bayesian approaches~\cite{Mackay1992}, including Monte Carlo Dropout~\citep{Mukhoti2020} and Deep Ensembles~\citep{Lakshminarayanan2017}, address this limitation by approximating posterior distributions over model parameters and thereby providing a principled framework for uncertainty estimation.
Although these methods often improve uncertainty calibration, their reliance on multiple stochastic forward passes introduces substantial computational overhead, which can be prohibitive for dense prediction tasks on high-resolution images.

Evidential Deep Learning (EDL)~\citep{Sensoy2018_edl} has emerged as a computationally efficient alternative for uncertainty-aware prediction. 
Rather than producing point estimates of class probabilities, EDL models the output as the parameters of a Dirichlet distribution and learns class-specific evidence values.
The amount of accumulated evidence directly reflects the confidence of the model, while a lack of evidence naturally translates into increased uncertainty. Consequently, EDL enables uncertainty estimation from a single deterministic forward pass and provides an attractive compromise between predictive performance and computational efficiency. Furthermore, the explicit representation of uncertainty makes EDL particularly suitable for OOD detection scenarios, where insufficient evidence should indicate unfamiliar observations.

Most existing EDL approaches rely on Euclidean objectives, such as expected mean squared error (MSE), for optimizing the evidential distribution.
While these objectives are analytically convenient, they do not explicitly account for the geometry of probability distributions and may encourage overconfident predictions. 
Recent work has therefore explored Wasserstein-based objectives as an alternative optimization criterion \citep{brosch}. In contrast to conventional point-wise losses, Wasserstein distances measure discrepancies directly in the space of probability distributions and respect the underlying geometry of the probability simplex. 
This property has been shown to improve distributional alignment and may lead to better calibrated uncertainty estimates, particularly in the presence of distributional shifts \citep{brosch}. The benefits of Wasserstein-based losses for uncertainty modeling have also been investigated in \citep{Xiong_10677853}.

This work investigates the influence of the Wasserstein order on uncertainty-aware semantic segmentation and OOD segmentation.
While first-order Wasserstein distances penalize discrepancies linearly, higher-order variants increasingly emphasize larger deviations between distributions. Consequently, different Wasserstein orders may induce different optimization dynamics and affect the balance between segmentation performance on in-distribution samples and uncertainty estimation for OOD observations.
This motivates a systematic investigation of first-, second-, and third-order Wasserstein objectives ($W_1$, $W_2$, and $W_3$)~\citep{peyre, Villani} within a unified evidential learning framework. Due to different specific numerical limitations of expected values we add different regularization parameters.
We add a MSE and Dice Loss term as well as KL-regularization.

In addition, uncertainty estimation is closely linked to the quality of the learned feature representation.
It therefore remains unclear whether the behavior of Wasserstein-based evidential learning is independent of the underlying network architecture or whether it interacts with specific representation-learning mechanisms. 
To address this question, we evaluate the proposed approach on two fundamentally different segmentation architectures: DeepLabV3+, representing convolutional neural networks, and SegFormer, representing transformer-based architectures. 
While DeepLabV3+ primarily exploits local spatial correlations through hierarchical convolutional feature extraction, SegFormer leverages self-attention mechanisms to capture long-range contextual dependencies.
Comparing both architectures provides insights into the relationship between feature representation, uncertainty estimation, and OOD robustness. 
The code is publicly available on \url{https://github.com/A-Brosch/EDL-OOD-Segmentation}.

The main contributions of this work are summarized as follows:

\begin{itemize}
\item We investigate Wasserstein-based EDL for uncertainty-aware semantic segmentation in the context of OOD detection.

\item We provide a systematic comparison of different Wasserstein orders ($W_1$, $W_2$, and $W_3$) and analyze their influence on segmentation performance, uncertainty calibration, and OOD segmentation capability.

\item We study the interaction between Wasserstein-based evidential learning and network architecture by evaluating both a convolutional backbone (DeepLabV3+) and a transformer-based backbone (SegFormer).

\item We conduct extensive experiments on established OOD segmentation benchmarks, including LostAndFound, RoadObstacle21, RoadAnomaly21, and Fishyscapes, and provide an empirical analysis of the resulting trade-offs between in-distribution accuracy and OOD awareness.
\end{itemize}

\section{Related Work}
Existing approaches for OOD segmentation can be broadly grouped into four categories: uncertainty-based methods, feature-space methods, approaches leveraging auxiliary models, and methods that incorporate additional OOD data during training.


\paragraph{Methods Leveraging Auxiliary OOD or Synthetic Data.}

One line of research uses additional real OOD data during training to explicitly expose the segmentation model to samples outside the closed set of in-distribution classes~\citep{Biase2021,Gao2023}. 
An early approach introduces an explicit void or anomaly class for supervision \citep{Bevandic}, while other works combine outlier exposure with dedicated loss functions and/or specialized network architectures. 
Examples include entropy maximization on OOD objects \citep{Chan2021}, energy-based abstention learning \citep{Tian2022}, hybrid discriminative-generative models \citep{Grcic2022}, and contrastive objectives that learn more robust anomaly representations \citep{Liu2023}. 
The paradigm is also extended to transformer-based mask-classification architectures, such as Mask2Anomaly \citep{Rai2023} and EAM \citep{Grcic2023}. 
Recent methods increasingly leverage foundation models and vision-language models to obtain more robust representations for OOD segmentation \citep{Holle2025,Marschall2025,Nayal2024,Zhong2024}.

To reduce the dependence on real OOD samples, several methods instead generate synthetic negative data during training. For example, normalizing flows have been employed to synthesize outlier samples that serve as auxiliary supervision for OOD segmentation \citep{Blum2019,Grcic2021,Gudovskiy2023,Won2026}. 
Moreover, RbA~\citep{Nayal2023} generates synthetic negatives and uses them to suppress the energy of OOD pixels, thereby improving the separation between in-distribution and OOD regions. 
UNO~\citep{Delic2024} combines negative-objectness scores obtained from synthetic outliers with uncertainty estimates to derive a more robust OOD score.

\paragraph{Auxiliary and Generative Reconstruction Models.}
The underlying assumption of a further group of methods is that regions from outside the training distribution are difficult to reconstruct or resynthesize faithfully. 
Besnier et al.~\citep{Besnier2021} operationalize this assumption by training a separate observer network that predicts, from adversarially perturbed features, where the primary segmentation network is likely to fail. 
A closely related strategy resynthesizes the input image from its predicted segmentation and flags pixels where the resynthesis diverges strongly from the original~\citep{Lis2019}; variants of this idea restrict the comparison to locally inpainted patches~\citep{Lis2020}, couple reconstruction and segmentation within a single joint model~\citep{Vojir2021}, or assess the reconstructability of individual image patches directly~\citep{Vojir2023}. 
Beyond image-level reconstruction, structure-constrained learning distills structural knowledge from a teacher network into a student model, improving segmentation of unseen classes across OOD, zero-shot, and domain-adaptation settings~\citep{Zhang2024}.

\paragraph{Feature-Space Approaches.}

Rather than relying on the network's predictive confidence, a third group of methods judges an input by where it falls within the learned feature space. 
cDNP~\citep{Galesso2023} and PixOOD~\citep{Vojir2024} estimate the in-distribution feature distribution non-parametrically or through class-conditional density models, treating pixels whose features fall outside the estimated support as anomalous. 
A related but distinct strategy fits an explicit density model with a normalizing flow: \citep{Blum2019} trains such a flow directly on the feature embeddings and scores pixels by their negative log-likelihood, and FlowEneDet~\citep{Gudovskiy2023} applies the same principle to the energy scores of an already frozen, pretrained segmentation model. 
\citep{Sodano2024} takes a complementary route, modeling each known class as a Gaussian while jointly learning a binary anomaly decoder that exploits similarity between classes, whereas Maskomaly~\citep{Ackermann2023} requires no additional training at all, repurposing the mask embeddings of a pretrained panoptic network and flagging anomalies through irregularities in their class-agnostic properties.

\paragraph{Uncertainty-Based Methods.}

A fourth direction relies purely on the network's own predictive behavior to flag OOD regions, without inspecting its internal features. 
The simplest such scores are read directly off the softmax output, most notably the maximum softmax probability~\citep{Hendrycks2016} and predictive entropy, both of which are cheap to compute but tend to stay overconfident on unfamiliar inputs \cite{Sensoy2018_edl}. 
Sampling-based alternatives instead aggregate several stochastic forward passes, for instance via Monte Carlo Dropout~\citep{Mukhoti2020} or Deep Ensembles~\citep{Lakshminarayanan2017}, trading substantially higher inference cost for improved calibration. 
A cheaper route to a similarly informative signal exploits the gradients of the loss landscape with respect to hypothesized labels, as in pixel-wise gradient norms (PGN, \citep{Maag2024_grads}), while Mahalanobis~\citep{Lee2018} and ODIN~\citep{Liang2018} instead perturb the input itself to sharpen the separation between in- and out-of-distribution confidence scores. 
Most closely related to our own work, Evidential Deep Learning~\citep{Sensoy2018_edl} departs from all of these by reformulating the output layer to directly parameterize a Dirichlet distribution over class probabilities, so that epistemic uncertainty can be read off from the accumulated evidence within a single deterministic forward pass~\citep{brosch}.

\paragraph{Positioning of this Work.}
In contrast to the approaches discussed above, our method requires neither auxiliary architectures, additional OOD training data, nor feature-space post-processing.  
Our approach belongs to the class of uncertainty-based OOD segmentation methods, as it derives uncertainty directly and deterministically from the evidential output of the segmentation network itself. 
Unlike sampling-based approaches such as MC Dropout or deep ensembles, our method requires only a single forward pass and does not rely on multiple stochastic predictions. Furthermore, in contrast to gradient-based or adversarial approaches, no additional backward passes or input perturbations are required. Therefore, our method is computationally closest to output-based uncertainty baselines such as maximum softmax probability.

Our own preliminary study established the general feasibility of Wasserstein-based evidential learning using a fixed first-order objective on a single convolutional architecture \citep{brosch}, and the present work extends this line of research along two axes. 
First, we systematically vary the order of the Wasserstein objective ($W_1$, $W_2$, $W_3$) to study its effect on the trade-off between in-distribution accuracy and OOD awareness. 
Second, we evaluate the resulting framework on both a convolutional architecture (DeepLabV3+) and a transformer-based architecture (SegFormer), which lets us disentangle properties of the objective from properties of the learned representation. 
In addition, we broaden the evaluation from LostAndFound, RoadObstacle21 and RoadAnomaly21 to the complete SegmentMeIfYouCan benchmark, including Fishyscapes.

\section{Method}

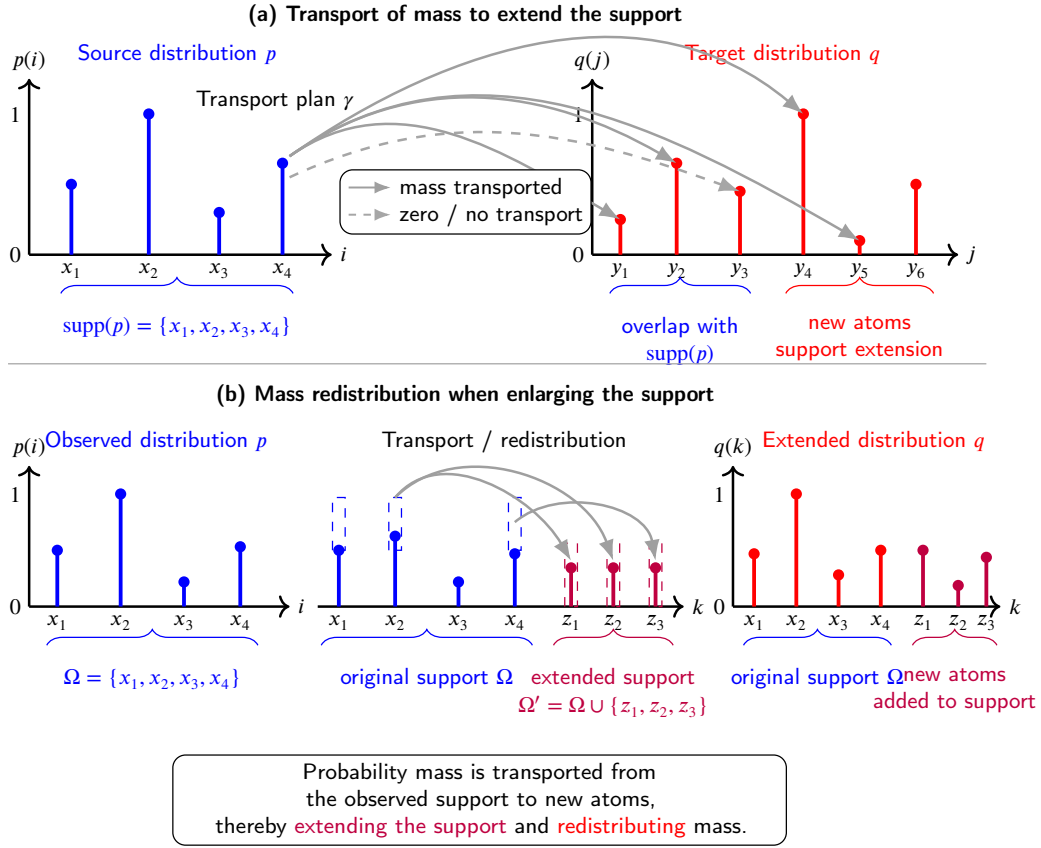
\begin{figure*}
\centering
\resizebox{0.8\textwidth}{!}{%
  \input{Images/Distribution_Transport}%
}
\caption{Illustration of probability mass transport for support extension.}
\label{fig:mass_transport_support_extension}
\end{figure*}

\paragraph{Background: Optimal Transport and Wasserstein Distances.}
Our work builds on the mathematical theory of optimal transport, which studies the minimal cost of transforming one probability distribution into another with respect to a given ground metric \citep{peyre,Villani}. 
From a measure-theoretic perspective, the training procedure of a segmentation network implicitly defines a probability measure over the joint space of image features and semantic labels, whose support is fixed by the finite collection of training instances and the label distribution observed during learning, so that the categories realized by this measure constitute the effective domain over which the network can make confident predictions. 
At deployment, however, the network is queried on inputs that may contain atoms outside this support, namely OOD objects that were never represented during training, so that the test distribution can be viewed as an extension of the training distribution by previously unseen atoms. This can be seen schematically in Fig.~\ref{fig:mass_transport_support_extension}.
Since any probability measure is normalized to unit total mass, accommodating these additional atoms is only possible by redistributing mass away from the atoms already present in the training support, and for an OOD input the network has no means of placing mass on the unseen atom directly, so this mismatch must instead be reflected in how the available mass is redistributed across the known atoms. 
This observation motivates casting OOD-aware prediction as a mass-transportation problem, for which optimal transport theory provides the natural mathematical language, characterizing the minimal cost of transforming one probability measure into another with respect to a chosen ground metric on the underlying space. 
Wasserstein distances instantiate this cost for probability distributions and thereby give a geometry-aware notion of discrepancy that, in contrast to divergences such as the KL divergence or cross-entropy, remains well-defined and informative even for distributions with disjoint or only partially overlapping support \citep{Villani,peyre}, precisely the regime in which OOD detection operates. 
This property has made Wasserstein distances attractive well beyond their origin in transportation theory. 
In deep learning, they have been popularized primarily through generative modeling, most notably the Wasserstein GAN \citep{ArjovskyCB17}, which replaces the Jensen-Shannon objective of the original GAN formulation with a Wasserstein critic to stabilize adversarial training, and through efficient approximate solvers such as entropic (Sinkhorn) regularization \citep{peyre}, which make optimal transport tractable at the scale required by deep networks. 
Closer to our setting, Wasserstein-based losses have also been proposed directly for supervised learning with structured or multi-label outputs \citep{Frogner2015}, exploiting a predefined ground metric between classes to obtain gradients that respect semantic similarity rather than treating all misclassifications as equally costly. It has been shown further to behave beneficial regarding the modeling of uncertainty shown in \citep{Xiong_10677853}. 
Our work differs from these applications in that we do not use the Wasserstein distance to compare a generator distribution to real data or to encode class-level semantic similarity, but instead employ it, within the evidential framework, as a geometry-aware alternative to the Euclidean (MSE) objective for training the parameters of a Dirichlet distribution, following the formulation recently proposed for evidential segmentation in our preliminary study \citep{brosch}, and directly motivated by the support-extension view of OOD detection developed above.

\paragraph{Evidential Deep Learning.}
EDL provides an alternative to both conventional softmax-based confidence estimation and computationally expensive Bayesian inference. 
Instead of interpreting network outputs as a single categorical probability vector, EDL models a higher-order distribution over class probabilities \cite{10.1007/978-3-031-31438-4_3,Sensoy2018_edl}. This allows the network to jointly represent class beliefs and epistemic uncertainty within a single deterministic forward pass.
Consequently, EDL is particularly attractive for dense prediction tasks such as semantic segmentation, where Bayesian sampling approaches often become computationally expensive.

In semantic segmentation, each pixel $z$ of an input image $x$ is assigned a class label $y$ from a predefined set of classes $i \in \{1,\ldots,C\}$. 

Given the corresponding logits $l_z$, the network obtains non-negative evidence $e_z$ by applying the ReLU activation element-wise,
\begin{equation}
    e_z = \operatorname{ReLU}(l_z) = \max(l_z,0) \enspace.
\end{equation}
These evidence values parameterize a Dirichlet distribution
\begin{equation}
S = \sum_{i=1}^{C} e_i + C = \sum_{i=1}^{C} \alpha_i \enspace,
\qquad
\alpha_i = e_i + 1 \enspace,
\end{equation}
where the Dirichlet distribution can be interpreted as a distribution over categorical probability distributions. The expected class probability $p_i$ and associated uncertainty $\mathcal{U}$ are given by
\begin{equation}
p_i = \frac{\alpha_i}{S} \enspace,
\qquad
\mathcal{U} = \frac{C}{S} \enspace.
\end{equation}
The quantity $S$ corresponds to the total accumulated evidence and therefore reflects the model's confidence in its prediction.
Large values of $S$ indicate strong support for a particular class assignment and consequently low uncertainty, whereas small values correspond to ambiguous observations or regions insufficiently represented by the training data. The quantity $S$ may further be interpreted as the concentration parameter of the Dirichlet distribution. Following the previous measure theoretic interpretation, a large $S$ corresponds to the class assigned to the pixel lies within the support of the learned in-training distribution. 
Conversely, small values result in flatter distributions that indicate insufficient evidence and increased epistemic uncertainty.
In this sense, EDL explicitly separates uncertainty arising from a lack of evidence from uncertainty caused by conflicting class beliefs.
Since OOD objects are not part of the training distribution, they are expected to generate little evidence for any known class, naturally resulting in high uncertainty.

Unlike softmax probabilities, which are normalized regardless of the quality of the underlying representation, evidential predictions explicitly distinguish between confident decisions supported by strong evidence and uncertain decisions characterized by insufficient evidence. This distinction is crucial for OOD segmentation, where the objective is not only to classify known semantic categories but also to identify regions for which the model lacks sufficient knowledge.

\paragraph{MSE Loss.}

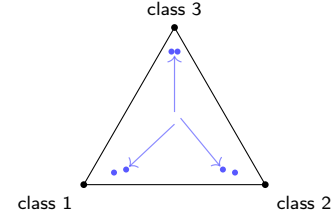
\begin{figure}
\centering
\input{Images/simplex_mse}
\caption{Illustration of the MSE effect on the probability simplex.}
\label{fig:simplex_mse}
\end{figure}

Most existing EDL formulations employ Euclidean objectives such as the expected mean squared error (MSE) loss
\begin{equation}
\mathcal{L}_{\text{MSE}_i}
=
(p_i-y_i)^2
+
\frac{p_i(1-p_i)}{S+1} \enspace,
\end{equation}
where $y_i$ denotes the one-hot encoded ground-truth label. The second term corresponds to the variance of the Dirichlet distribution and allows the model to jointly minimize prediction error and predictive uncertainty. 
While this objective generally yields sharp semantic boundaries and strong in-distribution segmentation performance, it also encourages predictions to move towards the vertices of the probability simplex, corresponding to highly confident one-hot decisions, see Fig.~\ref{fig:simplex_mse}.

Although such behavior is desirable for in-distribution classification, it is less suitable for OOD detection, where uncertainty should be preserved whenever insufficient evidence is available.
To address this limitation, we employ a Wasserstein-based objective that explicitly accounts for the geometry of probability distributions.

\paragraph{Wasserstein-based Loss.}
From a measure-theoretic perspective, the training procedure of a
segmentation network implicitly defines a probability measure
$\mu_{\text{train}}$ over the joint space of image features and semantic
labels, whose support is determined by the finite collection of atoms,
i.e., the training instances and the label distribution observed during
learning. The categories realized by $\mu_{\text{train}}$, together with
the range of visual appearances associated with each category, thus
constitute the effective domain over which the network can make confident
predictions. During deployment, however, the network is queried on samples
drawn from a distribution $\mu_{\text{test}}$ that may contain atoms outside
$\mathrm{supp}(\mu_{\text{train}})$, namely OOD objects that were never
represented during training. 
$\mu_{\text{test}}$ can therefore be regarded
as an extension of $\mu_{\text{train}}$ by previously unseen atoms.

Since any probability measure is normalized to unit total mass,
accommodating these additional atoms is only possible by redistributing
probability mass away from the atoms already present in
$\mathrm{supp}(\mu_{\text{train}})$. For an OOD input, however, the network
has no means of placing any mass on the corresponding unseen atom directly.
Instead, this mismatch must be reflected in how the available mass is
distributed across the known atoms. This observation motivates casting
OOD-aware prediction as a mass-transportation problem, for which optimal
transport theory~\citep{peyre, Villani} provides the natural mathematical
language: it characterizes the minimal cost of transforming one probability
measure into another with respect to a chosen ground metric on the
underlying space. Wasserstein distances instantiate this cost for
probability distributions and therefore provide a principled framework for
uncertainty-aware OOD detection. Rather than being forced, as under a
Euclidean objective, to concentrate the available mass on one of the known
atoms, the network is encouraged to withhold this assignment whenever the
accumulated evidence for the known atoms is insufficient to justify a
confident decision, corresponding to a low total evidence $S$ and hence a
high uncertainty $\mathcal{U} = C/S$. Since OOD objects are precisely those
inputs for which no known atom can absorb the assigned mass with sufficient
confidence, this construction directly connects the support-extension
argument above to the increased epistemic uncertainty that motivates the
remainder of this section.

The Wasserstein distance measures the minimal transport effort required to transform one probability distribution into another and therefore provides a geometry-aware notion of similarity. 
In contrast to Euclidean objectives, Wasserstein distances preserve information about the relative arrangement of probability mass and allow uncertainty to remain distributed across multiple classes when the available evidence is inconclusive. 

\begin{figure}
\centering
\input{Images/simplex_wasserstein}
\caption{Illustration of the Wasserstein effect on the probability simplex.}
\label{fig:simplex_wasserstein}
\end{figure}
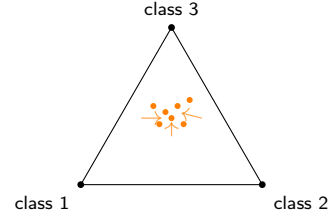

Using the Dirac metric as transport cost, the Wasserstein loss simplifies to
\begin{equation}
\mathcal{L}_{\mathcal{W}}
=
\mathbb{E}\left[1-p_y\right] \enspace,
\end{equation}
where $p_y$ denotes the predicted mean probability assigned to the ground-truth class. Note that using a unit transport cost corresponds to treating the support mismatch between distinct atoms as maximal.
This choice is motivated by safety-critical applications, where failing to correctly identify OOD objects may lead to hazardous downstream decisions. Consequently, all probability mass assigned to incorrect atoms is penalized equally, encouraging uncertainty rather than overconfident misclassification. Geometrically, this results in smoother transitions within the probability simplex and avoids the aggressive evidence concentration typically induced by MSE-based objectives (see Fig.~\ref{fig:simplex_wasserstein}).

A central contribution of this work is the investigation of the influence of the Wasserstein order on uncertainty-aware semantic segmentation. 
Previous Wasserstein-based EDL approaches \citep{brosch} predominantly employ first-order Wasserstein distances and do not analyze the effect of alternative orders. The general $q$-Wasserstein distance is defined as,
\begin{equation}
W_q(\mu,\nu)
=
\left(
\inf_{\gamma \in \Gamma(\mu,\nu)}
\int d(x,y)^q \, d\gamma(x,y)
\right)^{1/q} \enspace,
\end{equation}
where $\Gamma(\mu,\nu)$ denotes the set of admissible transport plans and $d(x,y)$ is the transport cost between two locations~\citep{peyre, Villani}.
The Wasserstein order determines the relative influence of large and small transport costs on the optimization objective.
For transport costs normalized to the interval $[0,1]$, increasing the order progressively suppresses the contribution of small discrepancies while relatively amplifying larger mismatches. Consequently, higher-order Wasserstein objectives focus increasingly on substantial distributional deviations.

First-order Wasserstein distances ($W_1$) weight transport costs linearly and therefore remain sensitive to both small and large discrepancies between distributions. Higher-order variants such as $W_2$ and $W_3$ increasingly emphasize larger deviations while suppressing the influence of smaller ones.

In the context of OOD segmentation, this property may have important implications. OOD samples frequently induce predictive distributions that differ substantially from those observed for in-distribution data. 
By assigning greater importance to such large discrepancies, higher-order Wasserstein objectives may improve the separation between in-distribution and OOD observations. 
Conversely, emphasizing large deviations too strongly may reduce sensitivity to subtle class-specific distinctions required for accurate semantic segmentation.
This suggests the existence of an order-dependent trade-off between segmentation accuracy and uncertainty awareness.

From a geometric viewpoint, lower-order Wasserstein objectives preserve local structure within the probability simplex and therefore maintain sensitivity to subtle class-specific variations. Higher-order objectives, on the other hand, increasingly prioritize global distributional discrepancies. 
Determining how these competing effects influence uncertainty estimation and OOD detection constitutes one of the central research questions of this work.

\paragraph{Regularization Terms.}
While the Wasserstein loss improves uncertainty calibration, it does not explicitly account for the spatial structure of segmentation outputs. 
To promote spatial consistency and accurate object boundaries, we therefore include a Dice loss term
\begin{equation}
\mathcal{L}_{\text{Dice}_i}
=
1-
\frac{2\sum_z p_i(z)\,y_i(z)}
{\sum_z p_i(z)+\sum_z y_i(z)} \enspace.
\end{equation}
Unlike pixel-wise objectives, the Dice loss operates on complete segmentation regions and therefore encourages coherent object shapes and improved overlap between predicted and ground-truth masks.

To further reduce overconfident predictions, particularly in ambiguous regions and OOD inputs, we employ a Kullback-Leibler regularization term
\begin{equation}
\mathcal{L}_{\text{KL}}
=
\mathrm{KL}
\left[
\mathrm{Dir}(\alpha)
\,\|\,
\mathrm{Dir}(1)
\right] \enspace,
\end{equation}
with Dir referring to the Dirichlet distribution.
This regularization encourages the predicted Dirichlet distribution to remain close to a non-informative uniform prior whenever insufficient evidence is available. In the evidential setting, this mechanism discourages unsupported evidence accumulation and prevents the model from assigning excessive confidence to observations for which little evidence exists. 
As a result, uncertain predictions are encouraged to collapse towards a vacuous prior distribution rather than towards arbitrary semantic classes, which is particularly beneficial in the presence of OOD objects. 
To avoid interfering with representation learning during the early training stages, the KL contribution is introduced gradually through an annealing schedule.

\paragraph{Loss Construction.}
The final training objective is defined as

\begin{equation}
\begin{aligned}
\mathcal{L}_{\text{total}}
={} &
\lambda_{\mathcal{W}}
\frac{1}{Z}
\sum_z \mathcal{L}_{\mathcal{W}}
+
\lambda_{\text{Dice}}
\frac{1}{C}
\sum_{i=1}^{C}
\mathcal{L}_{\text{Dice}_i} \\
&+
\lambda_{\text{KL}}
\frac{1}{Z}
\sum_z \mathcal{L}_{\text{KL}}
+
\lambda_{\text{MSE}}
\frac{1}{Z}
\sum_z
\sum_{i=1}^{C}
\mathcal{L}_{\text{MSE}_i} \enspace,
\label{eq:loss_total}
\end{aligned}
\end{equation}
where $Z$ denotes the number of pixels and $\lambda_{\mathcal{W}}$, $\lambda_{\text{Dice}}$, $\lambda_{\text{KL}}$, and $\lambda_{\text{MSE}}$ control the relative contribution of the individual components. 
Although the Wasserstein loss serves as the primary optimization objective, we retain a small MSE contribution throughout training. 
Empirically, pure Wasserstein optimization occasionally leads to slower convergence and reduced segmentation accuracy. 
The MSE term therefore acts as a stabilizing component that preserves sharp semantic boundaries while the Wasserstein objective maintains uncertainty awareness.
Overall, the proposed loss combines four complementary objectives: distributional alignment through Wasserstein optimization, structural consistency through Dice regularization, uncertainty calibration through KL regularization, and optimization stability through MSE supervision.

Although the proposed loss formulation is independent of the underlying segmentation architecture, the resulting uncertainty estimates may depend on the learned feature representations. 
We therefore evaluate the proposed framework on convolutional as well as on transformer-based architectures.
This comparison allows us to assess whether the behavior induced by different Wasserstein orders generalizes across fundamentally different representation-learning paradigms.

\section{Experiments}

\subsection{Experimental Setting}

\emph{Datasets.} \hspace{0.3ex}
For training, we employ the Cityscapes dataset~\cite{cordts2016}, one of the standard benchmarks for semantic segmentation in urban street scenes. Cityscapes consists of $2,975$ finely annotated training and $500$ validation images, 
captured in $50$ different European cities under varying weather and illumination conditions. 
The dataset provides high-resolution images ($1,\!024\times2,\!048$ pixels) with pixel-wise annotations for $19$ semantic classes, including roads, sidewalks, buildings, vehicles, pedestrians, and traffic infrastructure. Owing to its diversity and annotation quality, Cityscapes has become the de facto standard training dataset for semantic segmentation in autonomous driving and serves as the in-distribution (ID) data throughout this work.

For evaluating OOD segmentation, we follow the official SegmentMeIfYouCan (SMIYC) benchmark~\cite{Chan2021_smiyc}, which comprises four complementary datasets covering different OOD scenarios.
The \emph{LostAndFound (LAF)} dataset~\cite{Pinggera2016} contains $1,\!203$ real-world images recorded in urban driving environments with numerous small road obstacles, such as boxes, stones, tires, or plastic objects, deliberately placed on the roadway. 

\emph{RoadObstacle21}~\cite{Chan2021_smiyc} extends this setting by providing $412$ images with substantially greater diversity in both obstacle categories and environmental conditions. 
In contrast to LostAndFound, the dataset contains a wider range of everyday objects and naturally occurring road hazards, including construction material, debris, furniture, and other unexpected obstacles, thereby improving the evaluation of real-world robustness.

The Fishyscapes LostAndFound dataset~\citep{Blum2019} comprises $100$ validation images and $275$ non-public test images, with refined pixel-wise annotations distinguishing between OOD objects, in-distribution background corresponding to Cityscapes classes, and void regions containing all other pixels.

\emph{RoadAnomaly21}~\cite{Chan2021_smiyc} contains $100$ carefully curated images depicting highly diverse anomalous situations that are largely absent from conventional training datasets.
Rather than focusing solely on road obstacles, it includes unusual objects, animals, uncommon vehicles, and rare scene configurations, making it particularly suitable for evaluating the generalization capability of uncertainty-aware segmentation methods under severe distributional shifts.

\paragraph{Model Training.}
 
We evaluate the proposed loss formulation on both DeepLabV3+~\citep{Chen2018_deeplab} with a ResNet-50 backbone and SegFormer~\citep{xie2021segformersimpleefficientdesign}. 
DeepLabV3+ serves as a representative convolutional semantic segmentation architecture, which is commonly used in OOD segmentation literature \citep{Blum2019,brosch,Chan2021_smiyc}, while SegFormer represents a transformer-based approach with a hierarchical encoder and lightweight decoder. 
Comparing both architectures allows us to assess whether the proposed Wasserstein-based evidential learning generalizes consistently across different segmentation paradigms.

Training is conducted with a batch size of $4$ for $80$K iterations for DeepLabV3+ and $160$K iterations for SegFormer, using the Adam optimizer~\citep{kingma2015adam} with a learning rate of $3 \cdot 10^{-4}$ and a weight decay of $10^{-4}$. 
All experiments are performed on the Cityscapes training set and evaluated on the SegmentMeIfYouCan benchmark.

We run ablations on the weighting parameters $\lambda$ of our combined loss function, see \eqref{eq:loss_total}. The KL regularization weight is increased linearly from $0$ to $\lambda_{\text{KL}}$ between iterations $40$K and $48$K and then kept constant for the remainder of training. 
This annealing strategy allows the model to first learn discriminative semantic features while the learning rate is still relatively high and subsequently shifts the optimization focus towards uncertainty calibration and regularization. Without this scheduling, KL regularization tends to suppress evidence generation too early during training, resulting in reduced segmentation performance.

\paragraph{SegFormer-Specific Adaptation.}

When transferring the proposed loss formulation from DeepLabV3+ to SegFormer, we observed a fundamental difference in the logit distributions.
While DeepLabV3+ produces logits in the range of approximately $-550$ to $1500$, SegFormer logits are much more compressed, typically ranging from $-30$ to $5$. 
As a consequence, around 12\% of all pixels exhibit only negative logits. 
Since evidential deep learning derives the evidence values through a ReLU activation, these pixels receive zero evidence despite the model producing a valid segmentation prediction. 
This highlights the sensitivity of evidential learning to architecture-dependent logit calibration, an effect that has also been discussed for semantic segmentation models in \cite{Wang_2023_CVPR}. 
To mitigate this issue, we replace the ReLU activation with Softplus when evaluating SegFormer logits, ensuring that all predictions contribute non-zero evidence while preserving the ordering of the logits.
Compared to the standard Softplus function, 
\begin{equation}
\operatorname{softplus}(x) = \ln\!\left(1 + e^{x}\right) \enspace,
\end{equation}
we use the following numerically stable version,
\begin{equation}
\operatorname{softplus}(x)
=
\max(x,0)
+
\ln\!\left(1+e^{-|x|}\right) \enspace.
\end{equation}

\begin{figure*}
\centering
\includegraphics[width=0.45\linewidth]{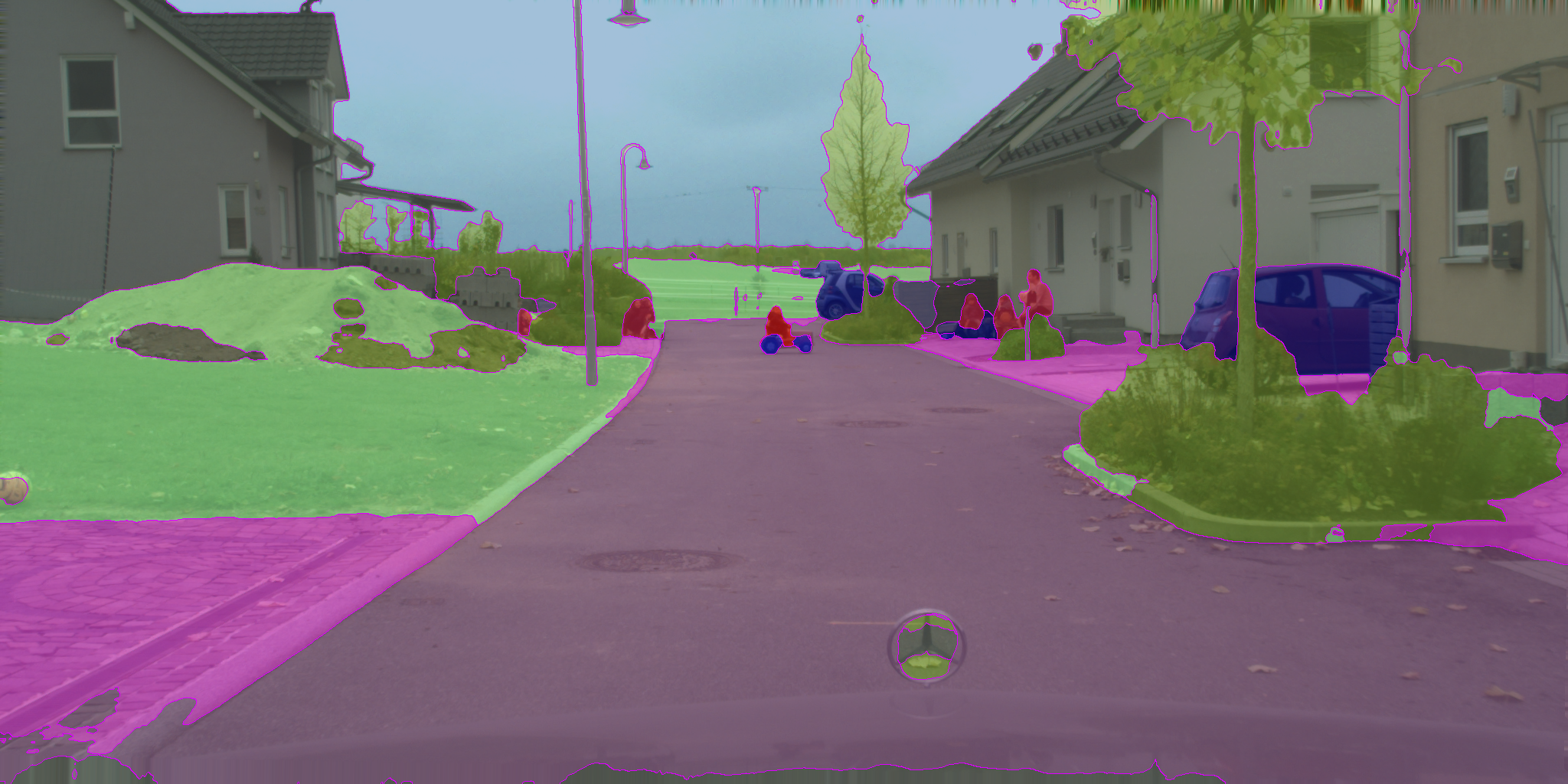}
\includegraphics[width=0.45\linewidth]{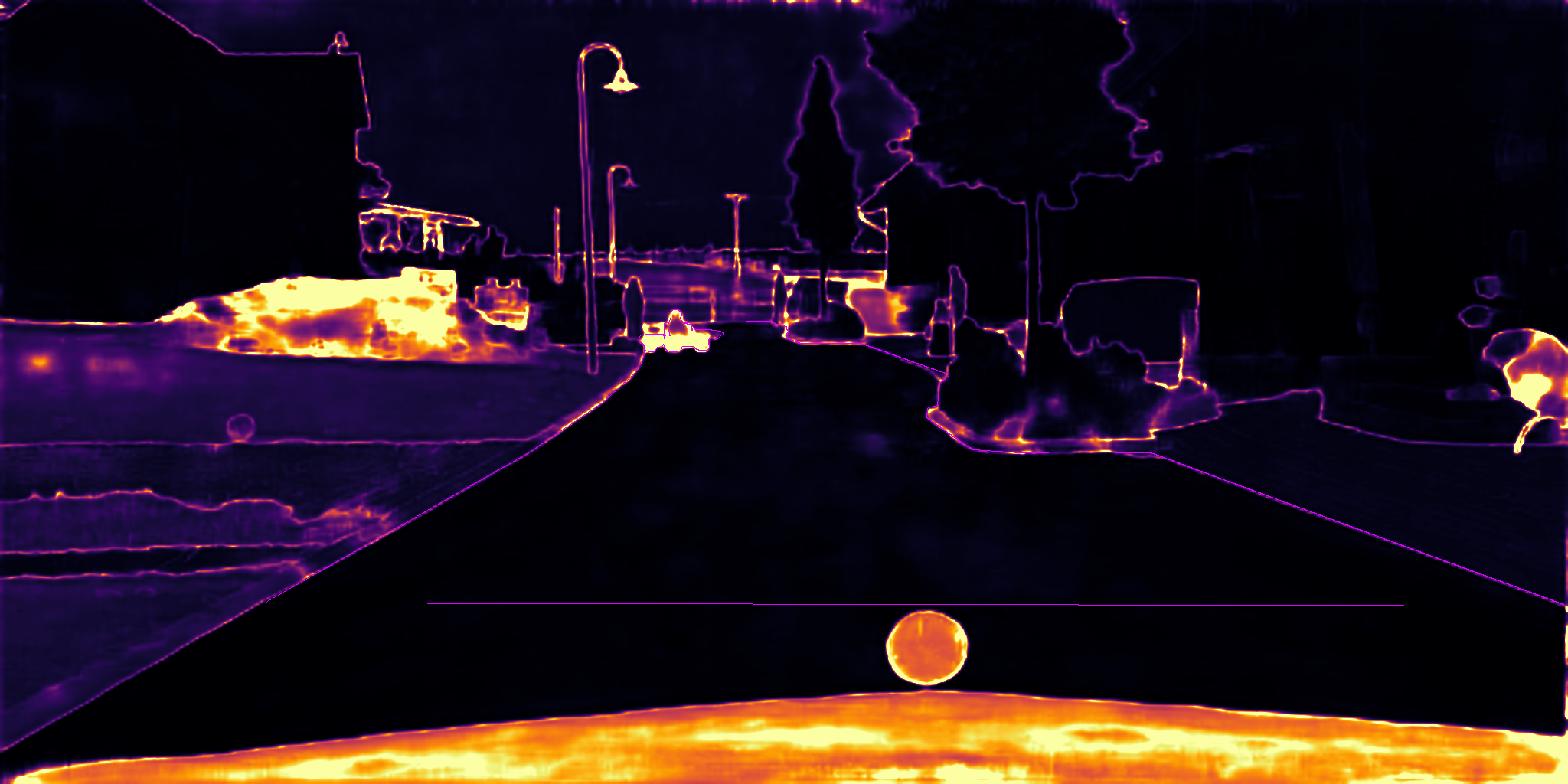} \\
\vspace{1.0ex}
\includegraphics[width=0.45\linewidth]{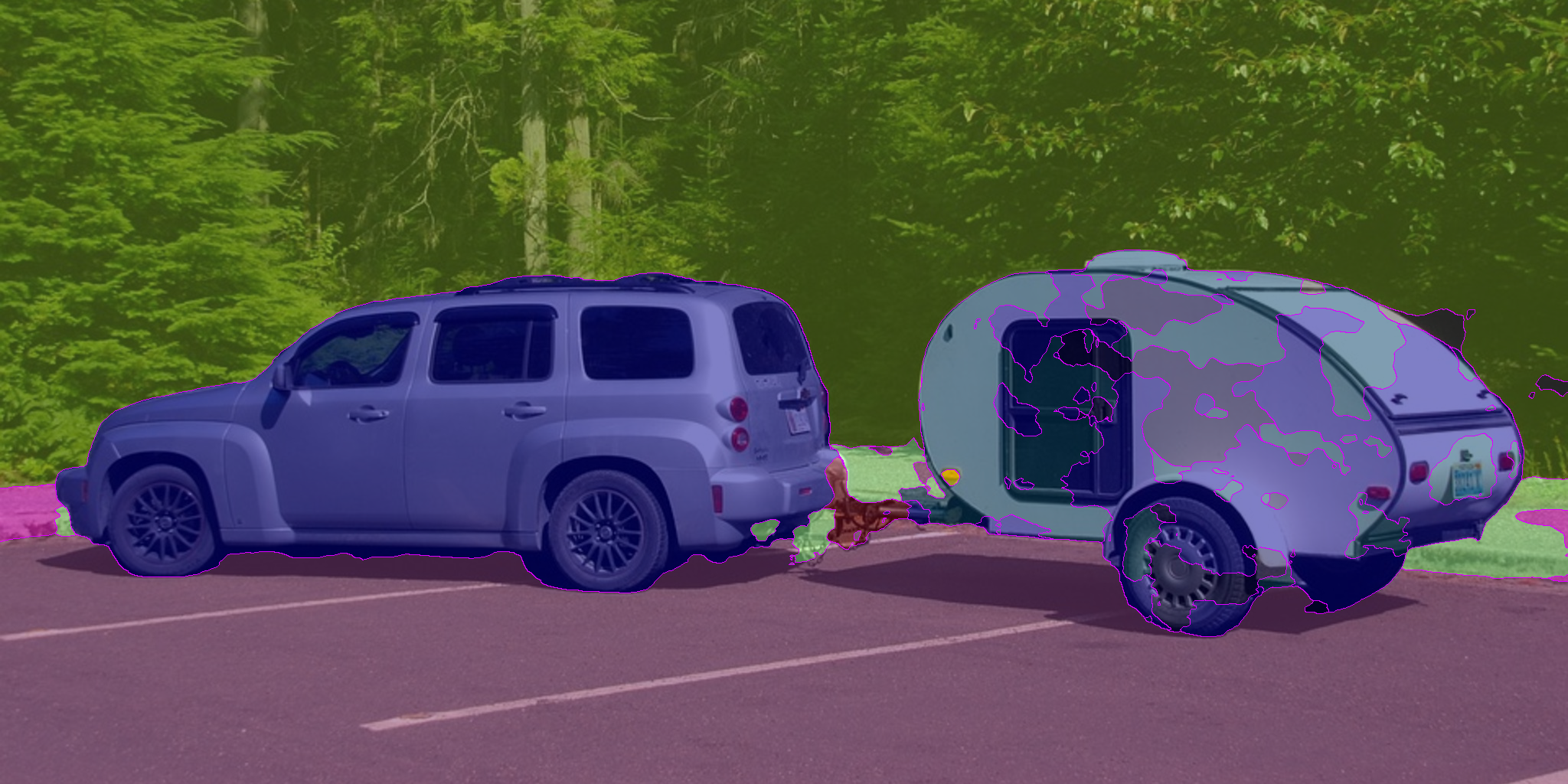}
\includegraphics[width=0.45\linewidth]{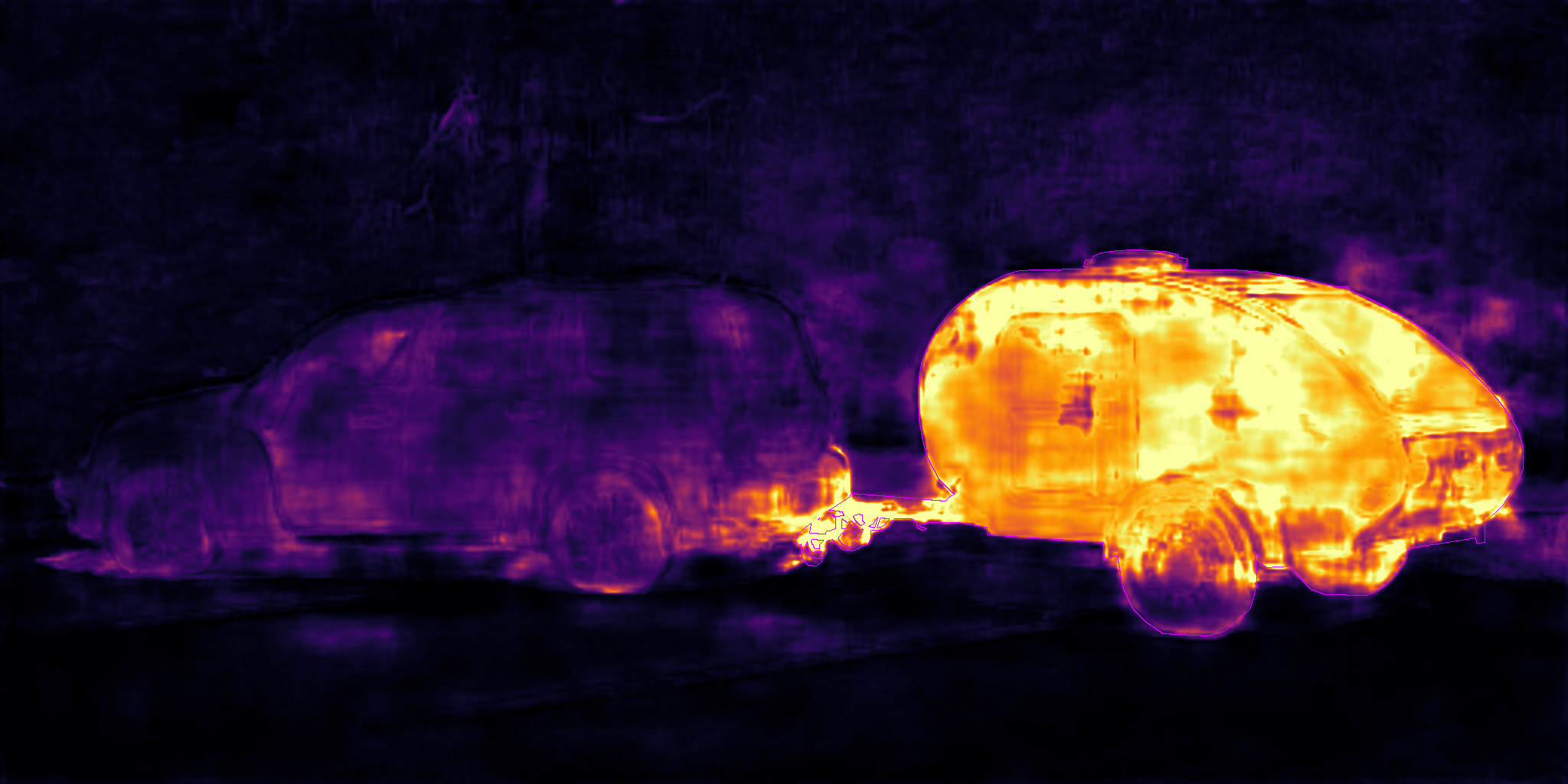}
\caption{ Left: Semantic segmentation prediction. Right: OOD heatmap obtained by our method using DeeplabV3+, $W_2$, $0.45$MSE and $0.75$Dice. The top images are from the LostAndFound dataset (a kid playing on the street next to a pile of rubble) and the bottom images from RoadAnomaly21 (a car with an attached caravan).}
\label{fig:examples}
\end{figure*}

\paragraph{Evaluation Metrics.}
To evaluate the OOD segmentation performance, we follow the official benchmark protocol and report both pixel-level and segment-level metrics. 
On the pixel-level, we use the area under the precision-recall curve (AuPRC) and the false positive rate at $95\%$ true positive rate (FPR$_{95}$). 
AuPRC measures the overall separability between OOD and in-distribution pixels across all thresholds, whereas FPR$_{95}$ quantifies the number of incorrectly detected OOD pixels when maintaining a high true positive detection rate. 
The latter is particularly relevant in safety-critical applications, where excessive false alarms may negatively affect downstream decision systems.

On the segment-level, we report the averaged adjusted similarity-weighted intersection-over-union (sIoU), positive predictive value (PPV), and $F_1$ score following the SegmentMeIfYouCan protocol. These metrics evaluate not only whether anomalous pixels are detected, but also whether complete anomalous objects are localized accurately. 
All segment-level measures are averaged across thresholds ranging from $0.25$ to $0.75$ with a step size of $0.05$.

Furthermore, standard semantic segmentation performance is assessed using mean Intersection-over-Union (mIoU) on Cityscapes.
Reporting both mIoU and OOD metrics allows us to analyze the trade-off between in-distribution segmentation performance and uncertainty-aware OOD detection, which constitutes one of the primary objectives of this work.

\subsection{Numerical Results}

We conduct a series of experiments to systematically evaluate the proposed loss method.
We first investigate different weightings of the individual loss components and analyze the influence of different Wasserstein orders ($W_1$, $W_2$, and $W_3$). 
Subsequently, we evaluate the proposed formulation across different segmentation architectures to assess its general applicability. 
The results are given in
Tables~\ref{tab:ood_laf_obstacle_dlv3_w1}--%
\ref{tab:ood_fishy_anomaly_segformer_w3},
and some qualitative examples are shown in Fig.~\ref{fig:examples}.
Finally, we compare the resulting models against established OOD segmentation baselines to quantify their performance in OOD segmentation.

\paragraph{Individual loss components.} 
We first analyze the individual loss components. 
Comparing pure MSE training ($\lambda_{\text{MSE}}=1$) against pure Wasserstein training ($\lambda_{\mathcal{W}}=1$), pure Wasserstein training generally provides stronger OOD performance than pure MSE training. 
Although this advantage is not consistent across all datasets and architectures, for example, in Fishyscapes LostAndFound, the MSE loss is often stronger than Wasserstein, whereas in the other three datasets, the opposite is observed. 
Both objectives generally perform considerably worse when optimized individually compared to their combination in the proposed Wasserstein MSE formulation, indicating that the two objectives complement each other. 
Using the DeepLabV3+ with $W_1$ configuration as a representative example (see Table~\ref{tab:ood_laf_obstacle_dlv3_w1}), adding this residual MSE contribution stabilizes training further. 
A weight of $\lambda_{\text{MSE}}=0.45$ gives the best overall trade off on LostAndFound and RoadObstacle21, improving AuPRC on LAF to $53.2$ over both the pure Wasserstein setting ($47.5$) and the pure MSE variant ($25.7$). We adopt this value as the reference for the remaining ablations in this configuration. 
Across the other Wasserstein orders and both architectures, an MSE weight between $0.40$ and $0.50$ remains consistently among the strongest choices, so this component generalizes well. 

Introducing the Dice term markedly improves FPR$_{95}$ for $W_1$, reducing it from $23.2$ at $\lambda_{\text{Dice}}=0$ to $17.6$ on LostAndFound at $\lambda_{\text{Dice}}=0.75$, while simultaneously raising $\overline{\text{sIoU}}$ and $\overline{F_1}$. 
Stronger Dice weighting ($\lambda_{\text{Dice}}=0.80$) reverses these gains and degrades performance across the board, confirming an optimal intermediate weighting for this order. 
The Dice regularization effect is less consistent across orders. For DeepLabV3+ with $W_2$ and $W_3$, increasing $\lambda_{\text{Dice}}$ does not reduce FPR$_{95}$ as clearly as for $W_1$, indicating that this particular benefit is order dependent rather than universal. 

Adding annealed KL regularization on top of the Wasserstein Dice MSE combination has a comparatively small and dataset dependent effect throughout our ablations. 
Consistent with this small and inconsistent effect, the KL term is not required for the best trade off: for DeepLabV3+ with $W_1$ order, the configuration $\lambda_{\mathcal{W}}=1.0$, $\lambda_{\text{MSE}}=0.45$, $\lambda_{\text{Dice}}=0.75$, $\lambda_{\text{KL}}=0.0$ already provides the most robust compromise on LostAndFound (AuPRC $47.4$, FPR$_{95}$ $17.6$, $\overline{\text{sIoU}}$ $26.6$, $\overline{\text{PPV}}$ $41.1$, $\overline{F_1}$ $24.2$), matching or exceeding the KL-regularized variants on the segmentation-quality metrics while attaining the lowest FPR$_{95}$ of all $W_1$ configurations. 
The KL term therefore primarily serves to calibrate the in-distribution predictions, as discussed below, rather than to improve OOD detection.

\paragraph{Wasserstein order.} 
Instead of a trade off that holds uniformly across all datasets, architectures, and metrics, we find that the preferred Wasserstein order depends primarily on the backbone and, to a lesser extent, on the dataset.

On DeepLabV3+, second-order objectives are favored on three of the four benchmarks.
$W_2$ achieves the strongest LostAndFound detection with $62.9$ AuPRC when combined with a Dice weight of $0.75$, and it also achieves the best RoadObstacle21 scores. On RoadAnomaly21, $W_2$ reaches $57.9$ AuPRC, clearly ahead of the best $W_3$ configuration at $43.7$ AuPRC. 
Only on Fishyscapes does $W_3$ achieve a higher AuPRC than $W_2$, and only by a small margin ($20.4$ compared to $18.0$). 
In general, $W_2$ is the preferred order on the convolutional backbone.
Third-order objectives on DeepLabV3+ are additionally penalized in terms of false-positive rate: FPR$_{95}$ remains above $80\%$ on both LostAndFound and RoadObstacle21 for the majority of $W_3$ configurations, compared to $17.6\%$ for the best $W_1$ or $W_2$ setting.

On SegFormer, the balance shifts toward the third order. 
$W_3$ attains the best AuPRC on all four benchmarks, reaching $47.9$ on LostAndFound, $42.1$ on RoadObstacle21, $38.2$ on Fishyscapes, and $64.5$ on RoadAnomaly21. On the obstacle datasets (i.e., LostAndFound and RoadObstacle21), this lead is pronounced ($W_3$ clearly ahead of both $W_1$ and $W_2$), whereas on the two more diverse anomaly datasets the advantage of $W_3$ over the best $W_2$ configuration is small ($64.5$ versus $59.6$ on RoadAnomaly21, $38.2$ versus $36.5$ on Fishyscapes). 
To summarize, rather than a single globally optimal order, our results point to an architecture-dependent optimum, i.e., $W_2$ is the preferred order for the convolutional backbone, with one exception on Fishyscapes, while $W_3$ is preferred for the transformer backbone, most clearly on the obstacle datasets.

\paragraph{Architecture comparison.} Beyond the choice of objective, we studied the interaction between Wasserstein-based evidential learning and the underlying network architecture by evaluating both a convolutional backbone (DeepLabV3+ with ResNet-50) and a transformer-based backbone (SegFormer).

The architecture comparison shows that the optimal configuration differs between backbones, particularly with respect to the Wasserstein order
DeepLabV3+ attains its best detection with $W_2$ and comparatively strong Dice regularization, whereas SegFormer reaches its strongest results with $W_3$ under light regularization. 
On RoadAnomaly21 the best SegFormer configuration is a lean $\lambda_{\text{MSE}}=0.40$ setup that reaches $64.5$ AuPRC, the highest OOD detection score of the entire study. 
This suggests that the globally aggregated features of the transformer interact differently with the emphasis on large distributional deviations, favoring higher-order objectives and requiring less residual MSE stabilization than the convolutional backbone. 
This backbone-dependent optimum is also reflected in a dataset specialization: SegFormer leads on three of the four OOD benchmarks, namely RoadObstacle21, Fishyscapes, and RoadAnomaly21, whereas DeepLabV3+ retains a clear advantage on LostAndFound, whose small, well-delineated on-road objects appear well matched to the local receptive fields of the CNN.

\begin{table*}[t]
\caption{OOD segmentation benchmark results for the LostAndFound and RoadObstacle21 dataset. }
\centering
\resizebox{0.98\linewidth}{!}{
\begin{tabular}{cccc cc ccc cc ccc}
\toprule 
\multicolumn{4}{c}{} & \multicolumn{5}{c}{LostAndFound test-NoKnown} & \multicolumn{5}{c}{RoadObstacle21} \\ 
\cmidrule(r){5-9} \cmidrule(r){10-14}
& & & & AuPRC $\uparrow$ & FPR$_{95}$ $\downarrow$ & $\overline{\text{sIoU}}$ $\uparrow$ & $\overline{\text{PPV}}$ $\uparrow$ & $\overline{F_1}$ $\uparrow$
& AuPRC $\uparrow$ & FPR$_{95}$ $\downarrow$ & $\overline{\text{sIoU}}$ $\uparrow$ & $\overline{\text{PPV}}$ $\uparrow$ & $\overline{F_1}$ $\uparrow$ \\
\midrule
\multicolumn{4}{l}{Ensemble}         & 2.9 & 82.0 & 6.7 & 7.6 & 2.7      & 1.1 & 77.2 & 8.6 & 4.7 & 1.3 \\
\multicolumn{4}{l}{MC Dropout}       & 36.8 & 35.6 & 17.4 & 34.7 & 13.0 & 4.9  & 50.3 & 5.5  & 5.8  & 1.1 \\
\multicolumn{4}{l}{Maximum Softmax}  & 30.1 & 33.2 & 14.2 & \textbf{62.2} & 10.3 & 15.7 & 16.6 & \underline{19.7} & 15.9 & 6.3 \\
\multicolumn{4}{l}{Entropy}          & 47.1 & 21.6 & 30.7 & 42.1 & 30.2 & \textbf{28.4} & 26.7 & 14.7 & \underline{20.5} & \textbf{9.7} \\
\midrule
\multicolumn{4}{l}{PGN}              & \textbf{69.3} & \textbf{9.8}  & \textbf{50.0} & 44.8 & \textbf{45.4} & 16.5 & 19.7 & 19.5 & 14.8 & 7.4 \\
\multicolumn{4}{l}{ODIN}             & 52.9 & 30.0 & \underline{39.8} & 49.3 & \underline{34.5} & \underline{22.1} & \underline{15.3} & \textbf{21.6} & 18.5 & \underline{9.4} \\
\multicolumn{4}{l}{Mahalanobis}      & 55.0 & \underline{12.9} & 33.8 & 31.7 & 22.1 & 20.9 & \textbf{13.1} & 13.5 & \textbf{21.8} & 4.7 \\
\midrule
\multicolumn{4}{l}{Ours }  & \underline{62.9} & 49.0 & 21.1 & \underline{54.5} & 27.1 & 7.1 & 53.9 & 7.4 & 8.9 & 2.8 \\ 
\bottomrule
\end{tabular} }
\label{tab:comparison_obstacle}
\end{table*}

\begin{table*}[t]
\caption{OOD segmentation benchmark results for the Fishyscapes LostAndFound and RoadAnomaly21 dataset.}
\centering
\resizebox{0.98\linewidth}{!}{
\begin{tabular}{cccc cc ccc cc ccc}
\toprule 
\multicolumn{4}{c}{} & \multicolumn{5}{c}{Fishyscapes LostAndFound} & \multicolumn{5}{c}{RoadAnomaly21} \\ 
\cmidrule(r){5-9} \cmidrule(r){10-14}
& & & & AuPRC $\uparrow$ & FPR$_{95}$ $\downarrow$ & $\overline{\text{sIoU}}$ $\uparrow$ & $\overline{\text{PPV}}$ $\uparrow$ & $\overline{F_1}$ $\uparrow$
& AuPRC $\uparrow$ & FPR$_{95}$ $\downarrow$ & $\overline{\text{sIoU}}$ $\uparrow$ & $\overline{\text{PPV}}$ $\uparrow$ & $\overline{F_1}$ $\uparrow$ \\
\midrule
\multicolumn{4}{l}{Ensemble}         & 0.3  & 90.4 & 3.1  & 1.1  & 0.4  & 17.7 & 91.1 & 16.4 & 20.8 & 3.4 \\
\multicolumn{4}{l}{MC Dropout}       & 14.4 & 47.8 & 4.8  & 18.1 & 4.3  & 28.9 & 69.5 & \underline{20.5} & 17.3 & 4.3 \\
\multicolumn{4}{l}{Maximum Softmax}  & 5.6  & 40.5 & 3.5  & 9.5  & 1.8  & 28.0 & 72.1 & 15.5 & 15.3 & 5.4 \\
\multicolumn{4}{l}{Entropy}          & 15.8 & \underline{36.1} & 7.5  & 16.3 & 8.6  & 30.0 & 73.0 & 17.8 & 15.6 & 5.1 \\
\midrule
\multicolumn{4}{l}{PGN}              & \underline{26.9} & 36.6 & \underline{14.8} & \textbf{29.6} & \underline{16.5} & \underline{42.8} & \underline{56.4} & \textbf{25.8} & \underline{21.8} & \underline{9.7} \\
\multicolumn{4}{l}{ODIN}             & 15.5 & 38.4 & 9.9  & 21.9 & 9.7  & 33.1 & 71.7 & 19.5 & 17.9 & 5.2 \\
\multicolumn{4}{l}{Mahalanobis}      & \textbf{32.9} & \textbf{8.7}  & \textbf{19.6} & \underline{29.4} & \textbf{19.2} & 20.0 & 87.0 & 14.8 & 10.2 & 2.7 \\
\midrule
\multicolumn{4}{l}{Ours}           & 18.0 & 44.3 & 3.3 & 23.5 & 5.1 & \textbf{57.9} & \textbf{51.7} & \textbf{25.8} & \textbf{32.0} & \textbf{12.4} \\ 
\bottomrule
\end{tabular} }
\label{tab:comparison_anomaly}
\end{table*}

\paragraph{Comparison against baselines.}
We compare our method against 7 uncertainty-based baselines, i.e., sampling-based (Ensemble, MC Dropout), output-based (Maximum Softmax, Entropy), and adversarial samples/gradient-based methods (PGN, ODIN, Mahalanobis). 
The results for LostAndFound and RoadObstacle21 are given in Table~\ref{tab:comparison_obstacle}, and for Fishyscapes and RoadAnomaly21 in Table~\ref{tab:comparison_anomaly}. 
For all datasets we report a single, fixed DeepLabV3+ configuration rather than a per-metric optimum, so that the comparison remains transparent and is not subject to post-hoc selection across the many logged configurations. 
Following our finding that $W_2$ is the strongest order for the convolutional backbone on the obstacle benchmarks and remains highly competitive on the anomaly benchmarks, we use the second-order configuration $\lambda_{\mathcal{W}_2}=1.0$, $\lambda_{\text{MSE}}=0.45$, $\lambda_{\text{Dice}}=0.75$, $\lambda_{\text{KL}}=0.0$.

While gradient- and adversarial-based methods achieve higher performance on RoadObstacle21 and Fishyscapes, our method reaches competitive results without requiring additional backward passes or input distortions. 
Compared to the uncertainty-based baselines, our approach outperforms most sampling-based methods while relying on a single forward pass rather than multiple stochastic samples. 
For instance, on RoadAnomaly21 our method surpasses the strongest baseline PGN in AuPRC (57.9 vs. 42.8) while requiring substantially lower computational overhead. This combination of strong anomaly-ranking performance and minimal inference cost makes the approach particularly well suited to safety-critical, resource-constrained deployment.

\begin{table}[t]
\caption{In-distribution evaluation on the Cityscapes dataset using mIoU.  The rows correspond to the three Wasserstein orders $W_1$, $W_2$, and $W_3$, and the two columns to the DeepLabV3+ and SegFormer backbones. $+a$ and $-b$ describe the size to the largest and smallest deviation respectively.}
\centering
\renewcommand{\arraystretch}{1.5}
\begin{tabular}{lcc}
\toprule
& DeepLabV3+ & SegFormer \\
\midrule
Baseline & $80.09$ & $84.00$\\
\midrule
$W_1$ & $67.67_{-15.55}^{+9.43}$ & $80.61_{-0.24}^{+0.27}$ \\
$W_2$ & $56.14_{-5.01}^{+6.53}$ & $80.60_{-0.44}^{+0.38}$ \\
$W_3$ & $66.88_{-11.04}^{+9.46}$ & $80.54_{-0.36}^{+0.20}$ \\
\bottomrule
\end{tabular}
\label{tab:miou_cityscapes}
\end{table}

\paragraph{In-distribution performance.}
Table~\ref{tab:miou_cityscapes} reports the in-distribution segmentation performance on the Cityscapes validation set, measured by the mean intersection over union (mIoU) over all $19$ evaluation classes. 
The top row (\emph{Baseline}) lists the corresponding cross-entropy models trained without any evidential objective ($80.09\%$ for DeepLabV3+ and $84.00\%$ for SegFormer). 
Each remaining entry is written as $\mu_{-a}^{+b}$, where $\mu$ is the mean mIoU averaged over all loss-weighting configurations ($\lambda_{\text{MSE}}$, $\lambda_{\text{Dice}}$, and $\lambda_{\text{KL}}$) evaluated for that order and architecture, and $-a$ and $+b$ give the largest downward and upward deviation from this mean across those configurations. 
The pair $(\mu-a,\ \mu+b)$ thus brackets the worst- and best-case in-distribution mIoU attained for the corresponding order, and the width $a+b$ quantifies how sensitive segmentation quality is to the loss weighting. 

The two architectures behave fundamentally differently. 
SegFormer is remarkably stable: the average mIoU stays within a narrow band of $80.54\%$--$80.61\%$ across all three orders, and every configuration remains above $80.09\%$, regardless of order or loss weighting, so that in-distribution accuracy is essentially preserved throughout the entire ablation. 
This indicates that the in-distribution accuracy of SegFormer is essentially decoupled from the specific choice of Wasserstein order and from the surrounding regularization. 
Thus, the substantial differences in OOD detection performance observed across SegFormer configurations are therefore obtained without any measurable cost to segmentation quality, making the backbone comparatively robust and straightforward to calibrate.

DeepLabV3+ exhibits different behavior. 
Average mIoU varies considerably across orders, from $67.67\%$ for $W_1$ down to $56.14\%$ for $W_2$ and back up to $66.88\%$ for $W_3$, an $11$ percentage point drop for the very order that we identified as the strongest choice for OOD detection on this backbone.
Within each order, the spread between the weakest and strongest loss-weighting configuration is an order of magnitude larger than for SegFormer, reaching $24.98$ percentage points for $W_1$ (range $[52.12\%, 77.10\%]$) and $20.50$ points for $W_3$ (range $[55.84\%, 76.34\%]$).
Even $W_2$, the most stable order on this backbone, still spans $11.53$ points ($[51.13\%, 62.66\%]$). 
This sensitivity indicates that DeepLabV3+ segmentation quality is tightly entangled with the choice of Dice, KL, and residual MSE weighting, so that a loss configuration selected for strong OOD detection can incur a substantial and difficult-to-predict penalty in in-distribution accuracy, in contrast to SegFormer, where the two objectives can effectively be tuned independently.

Relative to the cross-entropy baselines, evidential training does incur a cost in in-distribution accuracy, but its magnitude again depends strongly on the architecture. 
On SegFormer the reduction is modest and uniform: the mean mIoU drops from the baseline's $84.00\%$ to roughly $80.5$--$80.6\%$, a loss of about $3.4$ percentage points that is virtually independent of the Wasserstein order and the loss weighting. 
On DeepLabV3+ the average reduction is larger and highly configuration dependent, from a baseline of $80.09\%$ down to mean values between $56\%$ and $68\%$, although the best individual configurations per order remain much closer to the baseline (up to $77.10\%$ for $W_1$ and $76.34\%$ for $W_3$). 
Crucially, this in-distribution penalty is accompanied by a substantial gain in OOD performance: the same evidential models improve OOD detection over the corresponding cross-entropy and sampling-based baselines by large margins, most notably on the diverse anomaly benchmarks (Tables~\ref{tab:comparison_obstacle} and~\ref{tab:comparison_anomaly}). 
Thus, our method trades a comparatively small and, for SegFormer, nearly negligible decrease in in-distribution segmentation quality for a markedly stronger ability to flag OOD objects, which is precisely the trade off desired in safety-critical perception.

\section{Conclusion}
This work introduced a Wasserstein based evidential framework for OOD segmentation.
Interpreting OOD objects as missing support in the training distribution motivates a geometry aware loss that avoids the overconfidence encouraged by Euclidean MSE.
Our ablations show that Wasserstein, Dice, and annealed KL regularization are complementary, with Dice providing the clearest reduction in FPR$_{95}$.

The preferred Wasserstein order depends on the backbone. 
For DeepLabV3+, $W_2$ achieves the strongest overall detection on three of four benchmarks and is therefore the most reliable default. For SegFormer, replacing ReLU with softplus resolves missing evidence caused by predominantly negative logits. 
With this adaptation, SegFormer leads on RoadObstacle21, RoadAnomaly21, and Fishyscapes, while DeepLabV3+ remains stronger on LostAndFound. 
These results show that evidence activation and transport order should be selected jointly with the architecture.

Across LostAndFound and RoadObstacle21, the proposed method outperforms the evaluated deterministic and sampling based baselines on most comparisons, with particularly clear gains on RoadObstacle21.
It requires only a single deterministic forward pass and no auxiliary models, gradient access, or additional OOD training data.

The current Dirac transport cost assigns equal cost to all class confusions, and the analysis is limited to integer Wasserstein orders.
Future work can investigate learned semantic cost matrices, fractional orders, architecture specific evidence mappings, and extensions to open vocabulary and video segmentation.

\section*{Funding}
M. Felsberg was partially supported by the Swedish Research Council through the project 2022-04266 and by Linkoping University through the Center of Excellence AI4x. 
The funders had no role in the study design, data collection and
analysis, preparation of the manuscript, or decision to submit the work.

\section*{Data and code availability}
The source code and experimental configurations used in this study are
publicly available at
\url{https://github.com/A-Brosch/EDL-OOD-Segmentation}.
No new datasets were generated in this study. All datasets used for
training and evaluation are publicly available from their respective
benchmark providers and are cited in the manuscript.

\section*{Declaration of competing interest}
The authors declare that they have no known competing financial
interests or personal relationships that could have appeared to
influence the work reported in this paper.


\appendix
\section{More Numerical Results}

\setcounter{table}{0}
\renewcommand{\thetable}{A.\arabic{table}}
This section contains all the numerical results in tabular form for the various parameter configurations.

\clearpage

\begin{table*}[p]
\caption{OOD segmentation results for LostAndFound and RoadObstacle21 using DeepLabV3+ with Wasserstein order $W_1$.} 
\centering
\resizebox{0.98\linewidth}{!}{
\begin{tabular}{cccc cc ccc cc ccc}
\toprule 
\multicolumn{4}{c}{} & \multicolumn{5}{c}{LostAndFound test-NoKnown} & \multicolumn{5}{c}{RoadObstacle21} \\ 
\cmidrule(r){5-9} \cmidrule(r){10-14}
$\lambda_{\mathcal{W}1}$ & $\lambda_{\text{Dice}}$ & $\lambda_{\text{KL}}$ & $\lambda_{\text{MSE}}$ &  AuPRC $\uparrow$ & FPR$_{95}$ $\downarrow$ & $\overline{\text{sIoU}}$ $\uparrow$ & $\overline{\text{PPV}}$ $\uparrow$ & $\overline{F_1}$ $\uparrow$
& AuPRC $\uparrow$ & FPR$_{95}$ $\downarrow$ & $\overline{\text{sIoU}}$ $\uparrow$ & $\overline{\text{PPV}}$ $\uparrow$ & $\overline{F_1}$ $\uparrow$ \\
\midrule
0.00 & 0.00 & 0.00 & 1.00   & 25.7 & 56.8 & 25.9 & 27.7 & 14.6 & 1.6  & 62.1 & \textbf{18.5} & 5.6  & 2.0 \\ 
1.00 & 0.00 & 0.00 & 0.00   & 47.5 & 27.4 & 21.5 & 39.2 & 19.7 & 4.1  & 66.0 & 6.9  & 8.5  & 2.0 \\ 
\midrule
1.00 & 0.00 & 0.00 & 0.40   & 50.4 & 25.8 & 21.2 & 38.0 & 19.2 & \textbf{30.7} & 42.3 & 11.6 & 26.4 & \textbf{8.7} \\ 
1.00 & 0.00 & 0.00 & 0.45   & 53.2 & 23.2 & 25.3 & 37.7 & 20.7 & 20.2 & \textbf{31.2} & 11.0 & 18.3 & 5.8 \\ 
1.00 & 0.00 & 0.00 & 0.50   & 44.4 & 30.5 & 25.2 & \textbf{42.8} & 23.6 & 12.2 & 32.6 & 3.6  & \textbf{34.0} & 3.4 \\ 
\midrule
1.00 & 0.70 & 0.00 & 0.45 & 39.3 & 36.1 & 23.4 & 41.0 & 20.9 & 11.7 & 58.7 & 17.2 & 14.9 & 7.3\\ 
1.00 & 0.75 & 0.00 & 0.45 & 47.4 & \textbf{17.6} & \textbf{26.6} & 41.1 & \textbf{24.2} & 8.8 & 35.3 & 12.1 & 14.2 & 5.3\\ 
1.00 & 0.80 & 0.00 & 0.45 & 27.0 & 39.8 & 14.7 & 25.3 & 8.7 & 3.3 & 63.1 & 5.0 & 9.1 & 1.3\\ 
\midrule
1.00 & 0.75 & 0.10 & 0.45 & 42.1 & 25.5 & 19.8 & 36.2 & 16.3 & 2.8 & 51.1 & 10.4 & 7.4 & 2.7 \\ 
1.00 & 0.75 & 0.15 & 0.45 & \textbf{53.6} & 31.6 & 23.4 & 38.8 & 22.1 & 2.8 & 51.1 & 10.4 & 7.4 & 2.7 \\ 
1.00 & 0.75 & 0.20 & 0.45 & 49.1 & 38.2 & 21.9 & 37.5 & 20.1 & 4.9 & 39.0 & 16.2 & 5.4 & 2.7 \\ 
\bottomrule
\end{tabular} }
\label{tab:ood_laf_obstacle_dlv3_w1}
\end{table*}

\begin{table*}[p]
\caption{OOD segmentation results for LostAndFound and RoadObstacle21using DeepLabV3+ with Wasserstein order $W_2$. }
\centering
\resizebox{0.98\linewidth}{!}{
\begin{tabular}{cccc cc ccc cc ccc}
\toprule 
\multicolumn{4}{c}{} & \multicolumn{5}{c}{LostAndFound test-NoKnown} & \multicolumn{5}{c}{RoadObstacle21} \\ 
\cmidrule(r){5-9} \cmidrule(r){10-14}
$\lambda_{\mathcal{W}2}$ & $\lambda_{\text{Dice}}$ & $\lambda_{\text{KL}}$ & $\lambda_{\text{MSE}}$ &  AuPRC $\uparrow$ & FPR$_{95}$ $\downarrow$ & $\overline{\text{sIoU}}$ $\uparrow$ & $\overline{\text{PPV}}$ $\uparrow$ & $\overline{F_1}$ $\uparrow$
& AuPRC $\uparrow$ & FPR$_{95}$ $\downarrow$ & $\overline{\text{sIoU}}$ $\uparrow$ & $\overline{\text{PPV}}$ $\uparrow$ & $\overline{F_1}$ $\uparrow$ \\
\midrule
0.00 & 0.00 & 0.00 & 1.00   & 25.7 & 56.8 & \textbf{25.9} & 27.7 & 14.6 & 1.6  & 62.1 & \textbf{18.5} & 5.6  & 2.0 \\ 
1.00 & 0.00 & 0.00 & 0.00   & 43.8 & 38.9 & 18.9 & 53.9 & 21.6 & 1.1  & 44.5 & 14.1  & 3.6  & 1.7 \\ 
\midrule
1.00 & 0.00 & 0.00 & 0.40   & 50.4 & 36.9 & 18.2 & 45.3 & 20.3 & 13.7 & 53.9 & 8.0 & 15.2 & 3.3 \\ 
1.00 & 0.00 & 0.00 & 0.45   & 46.1 & 47.6 & 22.7 & 53.2 & 26.3 & 13.5 & 40.9 & 7.2 & 20.6 & 4.4 \\ 
1.00 & 0.00 & 0.00 & 0.45   & 48.8 & 38.9 & 17.0 & 44.4 & 18.4 & 17.2 & 49.9 & 6.4 & 13.7 & 3.0 \\ 
\midrule
1.00 & 0.70 & 0.00 & 0.45 & 47.5 & \textbf{35.0} & 23.5 & 44.4 & 25.9 & 13.5 & \textbf{30.7} & 11.2 & \textbf{22.0} & 7.0 \\ 
1.00 & 0.75 & 0.00 & 0.45 & \textbf{62.9} & 49.0 & 21.1 & \textbf{54.5} & \textbf{27.1} & 7.1 & 53.9 & 7.4 & 8.9 & 2.8 \\ 
1.00 & 0.80 & 0.00 & 0.45 & 27.4 & 52.1 & 12.9 & 35.2 & 12.6 & \textbf{18.7} & 63.0 & 16.4 & 20.4 & \textbf{9.2} \\ 
\midrule
1.00 & 0.75 & 0.10 & 0.50 & 19.5 & 47.5 & 10.8 & 26.7 & 8.8 & 2.4 & 95.9 & 8.8 & 4.0 & 1.8 \\ 
1.00 & 0.75 & 0.15 & 0.50 & 6.4 & 69.7 & 10.1 & 13.4 & 4.5 & 2.8 & 38.0 & 14.2 & 6.5 & 2.1 \\ 
1.00 & 0.75 & 0.20 & 0.50 & 36.5 & 51.8 & 14.1 & 40.0 & 14.0 & 1.4 & 65.5 & 9.6 & 5.1 & 2.0 \\ 
\bottomrule
\end{tabular} }
\label{tab:ood_laf_obstacle_dlv3_w2}
\end{table*}

\begin{table*}[p]
\caption{OOD segmentation results for LostAndFound and RoadObstacle21using DeepLabV3+ with Wasserstein order $W_3$.}
\centering
\resizebox{0.98\linewidth}{!}{
\begin{tabular}{cccc cc ccc cc ccc}
\toprule 
\multicolumn{4}{c}{} & \multicolumn{5}{c}{LostAndFound test-NoKnown} & \multicolumn{5}{c}{RoadObstacle21} \\ 
\cmidrule(r){5-9} \cmidrule(r){10-14}
$\lambda_{\mathcal{W}3}$ & $\lambda_{\text{Dice}}$ & $\lambda_{\text{KL}}$ & $\lambda_{\text{MSE}}$ &  AuPRC $\uparrow$ & FPR$_{95}$ $\downarrow$ & $\overline{\text{sIoU}}$ $\uparrow$ & $\overline{\text{PPV}}$ $\uparrow$ & $\overline{F_1}$ $\uparrow$
& AuPRC $\uparrow$ & FPR$_{95}$ $\downarrow$ & $\overline{\text{sIoU}}$ $\uparrow$ & $\overline{\text{PPV}}$ $\uparrow$ & $\overline{F_1}$ $\uparrow$ \\
\midrule
0.00 & 0.00 & 0.00 & 1.00   & 25.7 & \textbf{56.8} & \textbf{25.9} & 27.7 & 14.6 & 1.6  & 62.1 & \textbf{18.5} & 5.6  & 2.0 \\ 
1.00 & 0.00 & 0.00 & 0.00   & 34.0 & 88.1 & 13.8 & 40.1 & 17.0 & 12.0  & 74.1 & 8.4 & 18.6 & \textbf{5.2} \\ 
\midrule
1.00 & 0.00 & 0.00 & 0.40   & 10.5 & 93.2 & 1.3 & 12.0 & 1.2 & \textbf{13.7} & \textbf{50.6} & 4.7 & 19.3 & 3.8 \\ 
1.00 & 0.00 & 0.00 & 0.45   & 12.6 & 89.5 & 8.4 & 23.5 & 8.0 & 2.8 & 63.1 & 5.0 & 6.4 & 1.9 \\ 
1.00 & 0.00 & 0.00 & 0.50   & 11.7 & 84.9 & 2.1 & 10.6 & 1.6 & 5.4 & 73.7 & 2.9 & 6.0 & 1.4 \\ 
\midrule
1.00 & 0.70 & 0.00 & 0.45 & 15.5 & 86.2 & 4.3 & 16.9 & 3.4 & 2.4 & 82.6 & 3.4 & 5.3 & 0.7 \\ 
1.00 & 0.75 & 0.00 & 0.45 & 15.0 & 92.8 & 3.5 & 13.4 & 2.7 & 3.8 & 82.8 & 4.8 & 6.9 & 1.1 \\ 
1.00 & 0.80 & 0.00 & 0.45 & \textbf{46.3} & 82.7 & 13.9 & \textbf{42.8} & \textbf{17.1} & 1.4 & 59.8 & 4.7 & 4.1 & 1.0 \\ 
\midrule
1.00 & 0.75 & 0.10 & 0.45 & 9.5 & 93.4 & 2.6 & 15.1 & 1.6 & 1.4 & 90.1 & 3.5 & 2.5 & 0.3 \\ 
1.00 & 0.75 & 0.15 & 0.45 & 33.8 & 83.4 & 10.2 & 27.8 & 9.5 & 5.8 & 54.6 & 4.8 & \textbf{10.4} & 1.7 \\ 
1.00 & 0.75 & 0.20 & 0.45 & 7.0 & 89.0 & 6.2 & 14.4 & 4.0 & 7.4 & 81.5 & 5.4 & 9.2 & 1.9 \\ 
\bottomrule
\end{tabular} }
\label{tab:ood_laf_obstacle_dlv3_w3}
\end{table*}

\begin{table*}[p]
\caption{OOD segmentation results for Fishyscapes and RoadAnomaly21 using DeepLabV3+ with Wasserstein order $W_1$.}
\centering
\resizebox{0.98\linewidth}{!}{
\begin{tabular}{cccc cc ccc cc ccc}
\toprule 
\multicolumn{4}{c}{} & \multicolumn{5}{c}{Fishyscapes LostAndFound} & \multicolumn{5}{c}{RoadAnomaly21} \\ 
\cmidrule(r){5-9} \cmidrule(r){10-14}
$\lambda_{\mathcal{W}1}$ & $\lambda_{\text{Dice}}$ & $\lambda_{\text{KL}}$ & $\lambda_{\text{MSE}}$ &  AuPRC $\uparrow$ & FPR$_{95}$ $\downarrow$ & $\overline{\text{sIoU}}$ $\uparrow$ & $\overline{\text{PPV}}$ $\uparrow$ & $\overline{F_1}$ $\uparrow$
& AuPRC $\uparrow$ & FPR$_{95}$ $\downarrow$ & $\overline{\text{sIoU}}$ $\uparrow$ & $\overline{\text{PPV}}$ $\uparrow$ & $\overline{F_1}$ $\uparrow$ \\
\midrule
0.00 & 0.00 & 0.00 & 1.00   & 25.7 & 56.8 & 25.9 & 27.7 & 14.6 & 1.6  & 62.1 & \textbf{18.5} & 5.6  & 2.0 \\ 
1.00 & 0.00 & 0.00 & 0.00   & 47.5 & 27.4 & 21.5 & 39.2 & 19.7 & 4.1  & 66.0 & 6.9  & 8.5  & 2.0 \\ 
\midrule
1.00 & 0.00 & 0.00 & 0.40   & 50.4 & 25.8 & 21.2 & 38.0 & 19.2 & \textbf{30.7} & 42.3 & 11.6 & 26.4 & \textbf{8.7} \\ 
1.00 & 0.00 & 0.00 & 0.45   & 53.2 & 23.2 & 25.3 & 37.7 & 20.7 & 20.2 & \textbf{31.2} & 11.0 & 18.3 & 5.8 \\ 
1.00 & 0.00 & 0.00 & 0.50   & 44.4 & 30.5 & 25.2 & \textbf{42.8} & 23.6 & 12.2 & 32.6 & 3.6  & \textbf{34.0} & 3.4 \\ 
\midrule
1.00 & 0.70 & 0.00 & 0.45 & 39.3 & 36.1 & 23.4 & 41.0 & 20.9 & 11.7 & 58.7 & 17.2 & 14.9 & 7.3\\ 
1.00 & 0.75 & 0.00 & 0.45 & 47.4 & \textbf{17.6} & \textbf{26.6} & 41.1 & \textbf{24.2} & 8.8 & 35.3 & 12.1 & 14.2 & 5.3\\ 
1.00 & 0.80 & 0.00 & 0.45 & 27.0 & 39.8 & 14.7 & 25.3 & 8.7 & 3.3 & 63.1 & 5.0 & 9.1 & 1.3\\ 
\midrule
1.00 & 0.75 & 0.10 & 0.45 & 42.1 & 25.5 & 19.8 & 36.2 & 16.3 & 2.8 & 51.1 & 10.4 & 7.4 & 2.7 \\ 
1.00 & 0.75 & 0.15 & 0.45 & \textbf{53.6} & 31.6 & 23.4 & 38.8 & 22.1 & 2.8 & 51.1 & 10.4 & 7.4 & 2.7 \\ 
1.00 & 0.75 & 0.20 & 0.45 & 49.1 & 38.2 & 21.9 & 37.5 & 20.1 & 4.9 & 39.0 & 16.2 & 5.4 & 2.7 \\ 
\bottomrule
\end{tabular} }
\label{tab:ood_fishy_anomaly_dlv3_w1}
\end{table*}

\begin{table*}[p]
\caption{OOD segmentation results for Fishyscapes and RoadAnomaly21 using DeepLabV3+ with Wasserstein order $W_2$.}
\centering
\resizebox{0.98\linewidth}{!}{
\begin{tabular}{cccc cc ccc cc ccc}
\toprule 
\multicolumn{4}{c}{} & \multicolumn{5}{c}{Fishyscapes LostAndFound} & \multicolumn{5}{c}{RoadAnomaly21} \\ 
\cmidrule(r){5-9} \cmidrule(r){10-14}
$\lambda_{\mathcal{W}2}$ & $\lambda_{\text{Dice}}$ & $\lambda_{\text{KL}}$ & $\lambda_{\text{MSE}}$ &  AuPRC $\uparrow$ & FPR$_{95}$ $\downarrow$ & $\overline{\text{sIoU}}$ $\uparrow$ & $\overline{\text{PPV}}$ $\uparrow$ & $\overline{F_1}$ $\uparrow$
& AuPRC $\uparrow$ & FPR$_{95}$ $\downarrow$ & $\overline{\text{sIoU}}$ $\uparrow$ & $\overline{\text{PPV}}$ $\uparrow$ & $\overline{F_1}$ $\uparrow$ \\
\midrule
0.00 & 0.00 & 0.00 & 1.00   & \textbf{25.7} & 56.8 & \textbf{25.9} & \textbf{27.7} & \textbf{14.6} & 1.6  & 62.1 & 18.5 & 5.6  & 2.0 \\ 
1.00 & 0.00 & 0.00 & 0.00  & 4.6 & 53.6 & 7.1 & 12.6 & 9.2 & 33.0  & 60.3 & 21.1  & 19.5  & 5.0 \\ 
\midrule
1.00 & 0.00 & 0.00 & 0.40  & 10.1 & 38.4 & 4.6 & 20.8 & 4.2 & 50.3 & 57.9 & 20.0 & 22.6 & 5.6 \\ 
1.00 & 0.00 & 0.00 & 0.45   & 2.5 & 42.0 & 11.0 & 11.8 & 7.3 & 42.9 & 59.2 & 18.9 & 23.5 & 6.0 \\ 
1.00 & 0.00 & 0.00 & 0.50   & 3.9 & 46.5 & 8.5 & 10.4 & 6.9 & 35.6 & 65.2 & 23.4 & 20.3 & 6.8 \\ 
\midrule
1.00 & 0.70 & 0.00 & 0.45 & 3.4 & 44.7 & 12.0 & 16.3 & 8.7 & 54.6 & 49.4 & 20.5 & 28.5 & 8.6 \\ 
1.00 & 0.75 & 0.00 & 0.45 & 18.0 & 44.3 & 3.3 & 23.5 & 5.1 & \textbf{57.9} & 51.7 & \textbf{25.8} & \textbf{32.0} & \textbf{12.4} \\ 
1.00 & 0.80 & 0.00 & 0.45 & 3.6 & 40.6 & 12.2 & 16.9 & 9.8 & 46.6 & 58.4 & 20.9 & 31.0 & 7.9 \\ 
\midrule
1.00 & 0.75 & 0.10 & 0.45 & 3.0 & \textbf{37.4} & 13.7 & 10.4 & 7.9 & 33.6 & \textbf{75.3} & 13.3 & 24.3 & 5.2 \\ 
1.00 & 0.75 & 0.15 & 0.45 & 4.5 & 42.3 & 15.2 & 15.7 & 12.9 & 38.6 & 70.2 & 19.0 & 22.2 & 4.4 \\ 
1.00 & 0.75 & 0.20 & 0.45 & 3.5 & 39.2 & 7.4 & 9.6 & 5.4 & 48.8 & 58.9 & 16.9 & 25.9 & 5.8 \\ 
\bottomrule
\end{tabular} }
\label{tab:ood_fishy_anomaly_dlv3_w2}
\end{table*}

\begin{table*}[p]
\caption{OOD segmentation results for Fishyscapes and RoadAnomaly21 using DeepLabV3+ with Wasserstein order $W_3$.}
\centering
\resizebox{0.98\linewidth}{!}{
\begin{tabular}{cccc cc ccc cc ccc}
\toprule 
\multicolumn{4}{c}{} & \multicolumn{5}{c}{Fishyscapes LostAndFound} & \multicolumn{5}{c}{RoadAnomaly21} \\ 
\cmidrule(r){5-9} \cmidrule(r){10-14}
$\lambda_{\mathcal{W}3}$ & $\lambda_{\text{Dice}}$ & $\lambda_{\text{KL}}$ & $\lambda_{\text{MSE}}$ &  AuPRC $\uparrow$ & FPR$_{95}$ $\downarrow$ & $\overline{\text{sIoU}}$ $\uparrow$ & $\overline{\text{PPV}}$ $\uparrow$ & $\overline{F_1}$ $\uparrow$
& AuPRC $\uparrow$ & FPR$_{95}$ $\downarrow$ & $\overline{\text{sIoU}}$ $\uparrow$ & $\overline{\text{PPV}}$ $\uparrow$ & $\overline{F_1}$ $\uparrow$ \\
\midrule
0.00 & 0.00 & 0.00 & 1.00   & \textbf{25.7} & 56.8 & \textbf{25.9} & 27.7 & 14.6 & 1.6  & \textbf{62.1} & 18.5 & 5.6  & 2.0 \\ 
1.00 & 0.00 & 0.00 & 0.00  & 7.7 & 53.3 & 6.2 & 23.6 & 20.7 & 36.0  & 85.0 & 16.5  & 23.8  & 6.5 \\ 
\midrule
1.00 & 0.00 & 0.00 & 0.40   & 2.7 & 68.3 & 1.6 & 13.3 & 2.4 & 40.6 & 67.2 & 17.0 & \textbf{26.8} & 6.3 \\ 
1.00 & 0.00 & 0.00 & 0.45   & 2.2 & 54.6 & 8.9 & 9.7 & 7.8 & 42.3 & 76.8 & \textbf{20.7} & 25.5 & \textbf{7.4} \\ 
1.00 & 0.00 & 0.00 & 0.50   & 2.3 & 52.5 & 5.8 & 10.8 & 4.7 & 40.2 & 76.3 & 15.2 & 28.3 & 6.0 \\ 
\midrule
1.00 & 0.70 & 0.00 & 0.45 & 2.7 & 72.9 & 3.7 & 10.3 & 2.8 & \textbf{43.7} & 81.5 & 12.9 & 22.0 & 4.4 \\ 
1.00 & 0.75 & 0.00 & 0.45 & 4.8 & 77.6 & 2.5 & 16.4 & 3.4 & 42.0 & 71.2 & 11.8 & 21.9 & 3.5 \\ 
1.00 & 0.80 & 0.00 & 0.45 & \textbf{20.4} & 67.2 & 4.7 & \textbf{32.5} & 6.1 & 41.4 & 67.9 & 15.5 & 20.1 & 4.3 \\ 
\midrule
1.00 & 0.75 & 0.10 & 0.45 & 13.9 & 73.4 & 2.1 & 19.9 & 2.7 & 31.0 & 70.7 & 10.9 & 20.7 & 2.9 \\ 
1.00 & 0.75 & 0.15 & 0.45 & 12.1 & \textbf{46.4} & 4.0 & 22.8 & 5.0 & 41.1 & 68.8 & 14.5 & 22.1 & 4.4 \\ 
1.00 & 0.75 & 0.20 & 0.45 & 14.9 & 59.9 & 4.8 & 24.8 & 6.5 & 33.5 & 87.3 & 14.5 & 21.2 & 3.9 \\ 
\bottomrule
\end{tabular} }
\label{tab:ood_fishy_anomaly_dlv3_w3}
\end{table*}

\begin{table*}[p]
\caption{OOD segmentation results for LostAndFound and RoadObstacle21 using SegFormer with Wasserstein order $W_1$.}
\centering
\resizebox{0.98\linewidth}{!}{
\begin{tabular}{cccc cc ccc cc ccc}
\toprule 
\multicolumn{4}{c}{} & \multicolumn{5}{c}{LostAndFound test-NoKnown} & \multicolumn{5}{c}{RoadObstacle21} \\ 
\cmidrule(r){5-9} \cmidrule(r){10-14}
$\lambda_{\mathcal{W}1}$ & $\lambda_{\text{Dice}}$ & $\lambda_{\text{KL}}$ & $\lambda_{\text{MSE}}$ &  AuPRC $\uparrow$ & FPR$_{95}$ $\downarrow$ & $\overline{\text{sIoU}}$ $\uparrow$ & $\overline{\text{PPV}}$ $\uparrow$ & $\overline{F_1}$ $\uparrow$
& AuPRC $\uparrow$ & FPR$_{95}$ $\downarrow$ & $\overline{\text{sIoU}}$ $\uparrow$ & $\overline{\text{PPV}}$ $\uparrow$ & $\overline{F_1}$ $\uparrow$ \\
\midrule
0.00 & 0.00 & 0.00 & 1.00   & 22.4 & 63.1 & 7.0 & 13.1 & \textbf{26.9} & 21.2 & 64.7 & 4.2 & 55.7 & 3.9 \\ 
1.00 & 0.00 & 0.00 & 0.00   & 16.1 & 57.7 & 4.5 & 11.6 & 1.3 & 8.9 & 66.1 & 3.0 & 13.6 & 0.8 \\ 
\midrule
1.00 & 0.00 & 0.00 & 0.40   & \textbf{41.9} & 43.9 & \textbf{23.3} & 25.5 & 13.1 & 29.1 & 48.6 & 13.0 & 20.9 & 5.2 \\ 
1.00 & 0.00 & 0.00 & 0.45   & 34.1 & \textbf{39.8} & 22.4 & 19.2 & 9.7 & 31.5 & \textbf{35.8} & 9.8 & 28.6 & 5.8 \\ 
1.00 & 0.00 & 0.00 & 0.50   & 34.1 & 49.8 & 16.7 & 23.8 & 9.2 & 24.3 & 39.0 & 11.0 & 13.8 & 3.0 \\ 
\midrule
1.00 & 0.70 & 0.00 & 0.45 & 28.1 & 56.1 & 17.8 & 20.6 & 8.1 & 25.9 & 53.0 & 11.0 & 25.2 & 5.8 \\ 
1.00 & 0.75 & 0.00 & 0.45 & 10.3 & 71.6 & 4.3 & 10.6 & 1.0 & 14.8 & 70.3 & 5.8 & 17.1 & 2.1 \\ 
1.00 & 0.80 & 0.00 & 0.45 & 35.6 & 56.5 & 20.6 & \textbf{26.7} & 11.8 & 29.3 & 69.4 & 13.1 & 33.4 & 8.0 \\ 
\midrule
1.00 & 0.75 & 0.10 & 0.45 & 11.3 & 69.7 & 3.9 & 11.2 & 1.2 & 9.9 & 78.0 & 1.7 & \textbf{63.4} & 1.2 \\ 
1.00 & 0.75 & 0.15 & 0.45 & 31.2 & 70.2 & 15.2 & 25.7 & 9.4 & \textbf{36.3} & 40.4 & \textbf{16.9} & 28.7 & \textbf{9.2} \\ 
1.00 & 0.75 & 0.20 & 0.45 & 30.0 & 53.6 & 17.5 & 19.6 & 8.4 & 30.3 & 54.7 & 13.0 & 16.7 & 4.3 \\ 
\bottomrule
\end{tabular} }
\label{tab:ood_laf_obstacle_segformer_w1}
\end{table*}

\begin{table*}[p]
\caption{OOD segmentation results for LostAndFound and RoadObstacle21 using SegFormer with Wasserstein order $W_2$.}
\centering
\resizebox{0.98\linewidth}{!}{
\begin{tabular}{cccc cc ccc cc ccc}
\toprule 
\multicolumn{4}{c}{} & \multicolumn{5}{c}{LostAndFound test-NoKnown} & \multicolumn{5}{c}{RoadObstacle21} \\ 
\cmidrule(r){5-9} \cmidrule(r){10-14}
$\lambda_{\mathcal{W}2}$ & $\lambda_{\text{Dice}}$ & $\lambda_{\text{KL}}$ & $\lambda_{\text{MSE}}$ &  AuPRC $\uparrow$ & FPR$_{95}$ $\downarrow$ & $\overline{\text{sIoU}}$ $\uparrow$ & $\overline{\text{PPV}}$ $\uparrow$ & $\overline{F_1}$ $\uparrow$
& AuPRC $\uparrow$ & FPR$_{95}$ $\downarrow$ & $\overline{\text{sIoU}}$ $\uparrow$ & $\overline{\text{PPV}}$ $\uparrow$ & $\overline{F_1}$ $\uparrow$ \\
\midrule
0.00 & 0.00 & 0.00 & 1.00   & 22.4 & 63.1 & 7.0 & 13.1 & \textbf{26.9} & 21.2 & 64.7 & 4.2 & \textbf{55.7} & 3.9 \\ 
1.00 & 0.00 & 0.00 & 0.00   & 23.2 & 65.7 & 10.3 & 12.8 & 3.7 & 15.1 & \textbf{46.5} & 3.3 & 24.3 & 1.5 \\ 
\midrule
1.00 & 0.00 & 0.00 & 0.40   & 21.0 & 62.9 & 7.3 & 14.6 & 2.9 & 18.9 & 67.9 & 7.7 & 11.6 & 2.2 \\ 
1.00 & 0.00 & 0.00 & 0.45   & 8.8 & 79.9 & 3.2 & 8.1 & 0.9 & 9.4 & 81.0 & 3.0 & 40.9 & 2.7 \\ 
1.00 & 0.00 & 0.00 & 0.50   & \textbf{34.8} & 62.9 & 17.8 & 20.5 & 9.1 & \textbf{27.7} & 54.7 & 10.0 & 24.6 & 4.7 \\ 
\midrule
1.00 & 0.70 & 0.00 & 0.45 & 30.3 & \textbf{54.0} & 13.1 & \textbf{21.3} & 6.5 & 19.5 & 52.0 & 8.0 & 17.8 & 2.9 \\ 
1.00 & 0.75 & 0.00 & 0.45 & 32.1 & 54.2 & \textbf{20.4} & 21.2 & 9.7 & 11.7 & 73.8 & 5.2 & 17.4 & 1.7 \\ 
1.00 & 0.80 & 0.00 & 0.45 & 21.8 & 74.6 & 9.0 & 16.8 & 3.9 & 18.6 & 63.5 & 9.4 & 19.4 & 3.1 \\ 
\midrule
1.00 & 0.75 & 0.10 & 0.45 & 23.1 & 66.7 & 10.3 & 14.5 & 3.9 & 25.1 & 58.7 & \textbf{12.5} & 25.5 & \textbf{6.5} \\ 
1.00 & 0.75 & 0.15 & 0.45 & 11.6 & 65.4 & 3.4 & 7.3 & 0.9 & 9.5 & 47.1 & 1.8 & 53.2 & 1.2 \\ 
1.00 & 0.75 & 0.20 & 0.45 & 10.9 & 72.1 & 4.8 & 8.9 & 1.7 & 5.4 & 77.5 & 0.0 & 8.9 & 0.0 \\ 
\bottomrule
\end{tabular} }
\label{tab:ood_laf_obstacle_segformer_w2}
\end{table*}

\begin{table*}[p]
\caption{OOD segmentation results for LostAndFound and RoadObstacle21 using SegFormer with Wasserstein order $W_3$.}
\centering
\resizebox{0.98\linewidth}{!}{
\begin{tabular}{cccc cc ccc cc ccc}
\toprule 
\multicolumn{4}{c}{} & \multicolumn{5}{c}{LostAndFound test-NoKnown} & \multicolumn{5}{c}{RoadObstacle21} \\ 
\cmidrule(r){5-9} \cmidrule(r){10-14}
$\lambda_{\mathcal{W}3}$ & $\lambda_{\text{Dice}}$ & $\lambda_{\text{KL}}$ & $\lambda_{\text{MSE}}$ &  AuPRC $\uparrow$ & FPR$_{95}$ $\downarrow$ & $\overline{\text{sIoU}}$ $\uparrow$ & $\overline{\text{PPV}}$ $\uparrow$ & $\overline{F_1}$ $\uparrow$
& AuPRC $\uparrow$ & FPR$_{95}$ $\downarrow$ & $\overline{\text{sIoU}}$ $\uparrow$ & $\overline{\text{PPV}}$ $\uparrow$ & $\overline{F_1}$ $\uparrow$ \\
\midrule
0.00 & 0.00 & 0.00 & 1.00   & 22.4 & 63.1 & 7.0 & 13.1 & 26.9 & 21.2 & 64.7 & 4.2 & 55.7 & 3.9 \\ 
1.00 & 0.00 & 0.00 & 0.00   & 33.0 & 63.0 & 13.2 & 21.6 & 6.9 & 20.5 & 54.1 & 6.1 & 27.9 & 3.4 \\ 
\midrule
1.00 & 0.00 & 0.00 & 0.40   & 38.8 & 59.7 & 20.0 & 30.7 & 14.1 & \textbf{42.1} & 57.8 & \textbf{23.3} & 26.7 & \textbf{11.8} \\ 
1.00 & 0.00 & 0.00 & 0.45   & 21.1 & 64.2 & 10.1 & 14.5 & 4.1 & 20.9 & 64.6 & 10.0 & 14.8 & 3.1 \\ 
1.00 & 0.00 & 0.00 & 0.50   & 35.5 & \textbf{47.5} & 18.7 & 19.3 & 8.9 & 29.5 & \textbf{34.6} & 14.5 & 24.1 & 6.8 \\ 
\midrule
1.00 & 0.70 & 0.00 & 0.45 & 16.5 & 75.2 & 9.8 & 14.4 & 4.0 & 15.0 & 79.3 & 9.6 & 30.1 & 5.3 \\ 
1.00 & 0.75 & 0.00 & 0.45 & \textbf{47.9} & 52.3 & \textbf{27.8} & \textbf{34.2} & \textbf{19.7} & 26.9 & 49.8 & 18.4 & 18.6 & 6.1 \\ 
1.00 & 0.80 & 0.00 & 0.45 & 19.1 & 77.7 & 8.5 & 13.6 & 3.2 & 24.1 & 46.7 & 9.5 & 27.1 & 4.9 \\ 
\midrule
1.00 & 0.75 & 0.10 & 0.45 & 15.0 & 64.5 & 8.3 & 9.7 & 2.4 & 17.1 & 44.6 & 3.3 & \textbf{74.5} & 3.1 \\ 
1.00 & 0.75 & 0.15 & 0.45 & 18.4 & 69.2 & 5.0 & 10.9 & 1.8 & 16.4 & 73.4 & 4.4 & 29.8 & 2.8 \\ 
1.00 & 0.75 & 0.20 & 0.45 & 13.0 & 73.0 & 8.2 & 12.0 & 2.4 & 14.7 & 72.2 & 8.6 & 19.5 & 3.5 \\ 
\bottomrule
\end{tabular} }
\label{tab:ood_laf_obstacle_segformer_w3}
\end{table*}

\begin{table*}[p]
\caption{OOD segmentation results for Fishyscapes and RoadAnomaly21 using SegFormer with Wasserstein order $W_1$.}
\centering
\resizebox{0.98\linewidth}{!}{
\begin{tabular}{cccc cc ccc cc ccc}
\toprule 
\multicolumn{4}{c}{} & \multicolumn{5}{c}{Fishyscapes LostAndFound} & \multicolumn{5}{c}{RoadAnomaly21} \\ 
\cmidrule(r){5-9} \cmidrule(r){10-14}
$\lambda_{\mathcal{W}1}$ & $\lambda_{\text{Dice}}$ & $\lambda_{\text{KL}}$ & $\lambda_{\text{MSE}}$ &  AuPRC $\uparrow$ & FPR$_{95}$ $\downarrow$ & $\overline{\text{sIoU}}$ $\uparrow$ & $\overline{\text{PPV}}$ $\uparrow$ & $\overline{F_1}$ $\uparrow$
& AuPRC $\uparrow$ & FPR$_{95}$ $\downarrow$ & $\overline{\text{sIoU}}$ $\uparrow$ & $\overline{\text{PPV}}$ $\uparrow$ & $\overline{F_1}$ $\uparrow$ \\
\midrule
0.00 & 0.00 & 0.00 & 1.00   & 32.6 & 70.1 & 14.4 & \textbf{49.7} & 14.4 & 48.2  & 96.4 & 30.1 & 29.5  & 8.9 \\ 
1.00 & 0.00 & 0.00 & 0.00   & 28.9 & 67.1 & 8.9 & 40.3 & 11.7 & \textbf{49.4} & 86.6 & \textbf{32.2}  & 27.9  & \textbf{9.9} \\ 
\midrule
1.00 & 0.00 & 0.00 & 0.40   & 50.4 & 25.8 & 21.2 & 38.0 & 19.2 & 30.7 & 42.3 & 11.6 & 26.4 & 8.7 \\ 
1.00 & 0.00 & 0.00 & 0.45   & 53.2 & 23.2 & 25.3 & 37.7 & 20.7 & 20.2 & \textbf{31.2} & 11.0 & 18.3 & 5.8 \\ 
1.00 & 0.00 & 0.00 & 0.50   & 44.4 & 30.5 & 25.2 & 42.8 & 23.6 & 12.2 & 32.6 & 3.6  & \textbf{34.0} & 3.4 \\ 
\midrule
1.00 & 0.70 & 0.00 & 0.45 & 39.3 & 36.1 & 23.4 & 41.0 & 20.9 & 11.7 & 58.7 & 17.2 & 14.9 & 7.3\\ 
1.00 & 0.75 & 0.00 & 0.45 & 47.4 & \textbf{17.6} & \textbf{26.6} & 41.1 & \textbf{24.2} & 8.8 & 35.3 & 12.1 & 14.2 & 5.3\\ 
1.00 & 0.80 & 0.00 & 0.45 & 27.0 & 39.8 & 14.7 & 25.3 & 8.7 & 3.3 & 63.1 & 5.0 & 9.1 & 1.3\\ 
\midrule
1.00 & 0.75 & 0.10 & 0.45 & 42.1 & 25.5 & 19.8 & 36.2 & 16.3 & 2.8 & 51.1 & 10.4 & 7.4 & 2.7 \\ 
1.00 & 0.75 & 0.15 & 0.45 & \textbf{53.6} & 31.6 & 23.4 & 38.8 & 22.1 & 2.8 & 51.1 & 10.4 & 7.4 & 2.7 \\ 
1.00 & 0.75 & 0.20 & 0.45 & 49.1 & 38.2 & 21.9 & 37.5 & 20.1 & 4.9 & 39.0 & 16.2 & 5.4 & 2.7 \\ 
\bottomrule
\end{tabular} }
\label{tab:ood_fishy_anomaly_segformer_w1}
\end{table*}

\begin{table*}[p]
\caption{OOD segmentation results for Fishyscapes and RoadAnomaly21 using SegFormer with Wasserstein order $W_2$.}
\centering
\resizebox{0.98\linewidth}{!}{
\begin{tabular}{cccc cc ccc cc ccc}
\toprule 
\multicolumn{4}{c}{} & \multicolumn{5}{c}{Fishyscapes LostAndFound} & \multicolumn{5}{c}{RoadAnomaly21} \\ 
\cmidrule(r){5-9} \cmidrule(r){10-14}
$\lambda_{\mathcal{W}2}$ & $\lambda_{\text{Dice}}$ & $\lambda_{\text{KL}}$ & $\lambda_{\text{MSE}}$ &  AuPRC $\uparrow$ & FPR$_{95}$ $\downarrow$ & $\overline{\text{sIoU}}$ $\uparrow$ & $\overline{\text{PPV}}$ $\uparrow$ & $\overline{F_1}$ $\uparrow$
& AuPRC $\uparrow$ & FPR$_{95}$ $\downarrow$ & $\overline{\text{sIoU}}$ $\uparrow$ & $\overline{\text{PPV}}$ $\uparrow$ & $\overline{F_1}$ $\uparrow$ \\
\midrule
0.00 & 0.00 & 0.00 & 1.00   & 32.6 & 70.1 & 14.4 & 49.7 & \textbf{14.4} & 48.2  & 96.4 & 30.1 & \textbf{29.5}  & 8.9 \\ 
1.00 & 0.00 & 0.00 & 0.00   & 28.2 & 72.2 & 6.8 & 32.0 & 8.1 & 48.9 & 85.1 & 33.5  & 24.4  & 8.1 \\ 
\midrule
1.00 & 0.00 & 0.00 & 0.40   & 35.6 & \textbf{55.0} & 6.9 & 63.3 & 11.5 & 50.3 & 87.3 & 31.0 & 26.9 & 8.6 \\ 
1.00 & 0.00 & 0.00 & 0.45   & 16.2 & 73.0 & 2.7 & 31.0 & 3.5 & 44.6 & \textbf{75.0} & \textbf{35.3} & 26.0 & \textbf{10.6} \\ 
1.00 & 0.00 & 0.00 & 0.50   & \textbf{36.5} & 70.5 & 8.2 & 53.7 & \textbf{11.9} & \textbf{59.6} & 79.4 & 37.8 & 28.4 & 12.0 \\ 
\midrule
1.00 & 0.70 & 0.00 & 0.45 & 30.9 & 64.5 & 7.5 & 48.6 & 9.9 & 50.6 & 88.3 & 32.5 & 24.6 & 8.6 \\ 
1.00 & 0.75 & 0.00 & 0.45 & 31.8 & 68.1 & 10.2 & 49.2 & \textbf{14.3} & 55.0 & 79.9 & 33.9 & 28.4 & 10.0 \\ 
1.00 & 0.80 & 0.00 & 0.45 & \textbf{36.5} & 65.1 & 7.8 & 49.2 & 10.3 & 49.9 & 86.0 & 30.4 & 28.6 & 8.6 \\ 
\midrule
1.00 & 0.75 & 0.10 & 0.45 & 29.9 & 73.7 & 7.9 & 47.8 & 11.1 & 48.3 & 93.1 & 30.0 & 27.5 & 8.1 \\ 
1.00 & 0.75 & 0.15 & 0.45 & 24.4 & 65.2 & 5.3 & 32.7 & 5.9 & 45.7 & 88.2 & 30.6 & 24.8 & 7.7 \\ 
1.00 & 0.75 & 0.20 & 0.45 & 13.5 & 62.1 & 3.7 & 41.1 & 5.1 & 48.0 & 80.6 & 29.1 & 28.6 & 8.4 \\ 
\bottomrule
\end{tabular} }
\label{tab:ood_fishy_anomaly_segformer_w2}
\end{table*}

\begin{table*}[p]
\caption{OOD segmentation results for Fishyscapes and RoadAnomaly21 using SegFormer with Wasserstein order $W_3$.}
\centering
\resizebox{0.98\linewidth}{!}{
\begin{tabular}{cccc cc ccc cc ccc}
\toprule 
\multicolumn{4}{c}{} & \multicolumn{5}{c}{Fishyscapes LostAndFound} & \multicolumn{5}{c}{RoadAnomaly21} \\ 
\cmidrule(r){5-9} \cmidrule(r){10-14}
$\lambda_{\mathcal{W}3}$ & $\lambda_{\text{Dice}}$ & $\lambda_{\text{KL}}$ & $\lambda_{\text{MSE}}$ &  AuPRC $\uparrow$ & FPR$_{95}$ $\downarrow$ & $\overline{\text{sIoU}}$ $\uparrow$ & $\overline{\text{PPV}}$ $\uparrow$ & $\overline{F_1}$ $\uparrow$
& AuPRC $\uparrow$ & FPR$_{95}$ $\downarrow$ & $\overline{\text{sIoU}}$ $\uparrow$ & $\overline{\text{PPV}}$ $\uparrow$ & $\overline{F_1}$ $\uparrow$ \\
\midrule
0.00 & 0.00 & 0.00 & 1.00   & 32.6 & 70.1 & 14.4 & 49.7 & \textbf{14.4} & 48.2  & 96.4 & 30.1 & 29.5  & 8.9 \\ 
1.00 & 0.00 & 0.00 & 0.00  & 28.4 & 76.0 & 9.1 & 49.1 & 12.2 & 57.5 & 84.4 & 34.3  & 28.9  & 10.6 \\ 
\midrule
1.00 & 0.00 & 0.00 & 0.40   & 37.3 & 71.2 & 12.7 & 54.4 & \textbf{18.8} & \textbf{64.5} & \textbf{72.4} & 35.7 & 31.7 & 13.6 \\ 
1.00 & 0.00 & 0.00 & 0.45   & 22.4 & 69.0 & 8.4 & 58.5 & 13.6 & 42.4 & 90.1 & 30.3 & 26.3 & 8.3 \\ 
1.00 & 0.00 & 0.00 & 0.50   & 38.0 & \textbf{60.3} & 10.3 & \textbf{66.7} & 15.5 & 56.8 & 85.0 & 36.6 & 28.6 & 11.4 \\ 
\midrule
1.00 & 0.70 & 0.00 & 0.45 & 24.1 & 73.0 & 5.4 & 50.8 & 7.4 & 46.4 & 82.8 & 31.8 & 25.7 & 7.7 \\ 
1.00 & 0.75 & 0.00 & 0.45 & \textbf{38.2} & 73.3 & 11.2 & 43.5 & 15.7 & 63.4 & 74.6 & \textbf{36.9} & \textbf{32.8} & \textbf{13.8} \\ 
1.00 & 0.80 & 0.00 & 0.45 & 21.8 & 78.7 & 9.6 & 35.9 & 10.5 & 42.1 & 93.2 & 30.3 & 24.7 & 6.9 \\ 
\midrule
1.00 & 0.75 & 0.10 & 0.45 & 26.6 & 64.5 & 8.8 & 49.6 & 11.4 & 49.4 & 83.1 & 32.8 & 27.3 & 8.7 \\ 
1.00 & 0.75 & 0.15 & 0.45 & 34.5 & 69.9 & 9.9 & 43.9 & 13.7 & 46.4 & 93.2 & 32.4 & 25.8 & 8.1 \\ 
1.00 & 0.75 & 0.20 & 0.45 & 17.4 & 69.0 & 4.6 & 39.9 & 5.9 & 42.3 & 94.2 & 32.2 & 28.5 & 8.9 \\ 
\bottomrule
\end{tabular} }
\label{tab:ood_fishy_anomaly_segformer_w3}
\end{table*}


\clearpage
\printcredits


\bibliographystyle{cas-model2-names}
\bibliography{biblio}


\bio{}
ARNOLD BROSCH received the M.Sc. in Mathematics in $2024$ from Bergische Universit\"at Wuppertal. He is currently a PhD student at Heinrich-Heine-University D\"usseldorf. 
His research interests include computer vision, uncertainty quantification, robustness in deep learning, representation learning, optimal transport, stochastic analysis.
\endbio

\bio{}
ABDELRAHMAN ELDESOKEY received the Ph.D. degree in computer vision and deep learning from Link\"oping University, Sweden, in 2021, and was a Postdoctoral Fellow with King Abdullah University of Science and Technology (KAUST), Saudi Arabia. He is currently an Assistant Professor with Hamad Bin Khalifa University (HBKU), Qatar, where he leads the Vision, Generative AI and Perception (VGAP) Lab. 
His research spans computer vision, controllable image and video generation, multi-modal large language models and agentic systems, and 3D scene understanding.
\endbio

\bio{}
MICHAEL FELSBERG received the Ph.D. degree from Kiel University, Germany, in 2002, and the Docent degree from Link\"oping University, Sweden, in 2005. He has been a Full Professor with Link\"oping University, since 2008. His research focuses on visual learning theory and includes also video object and instance segmentation, classification, segmentation, and registration of point clouds, as well as efficient machine learning techniques for incremental, few-shot, and long-tailed settings. He received the DAGM Olympus Award in 2005, the PAMI Mark Everingham Prize in 2026, and is a fellow of IAPR, ELLIS, and AAIA.
\endbio

\bio{}
KIRA MAAG received the Ph.D. degree in Mathematics from the University of Wuppertal, Germany, in 2021. 
Since 2024, she has been a Junior Professor of Visual Analytics in Computer Science with Heinrich Heine University D\"usseldorf, Germany.
Her research focuses on trustworthy artificial intelligence for computer vision, including uncertainty quantification, robustness, adversarial attacks, out-of-distribution detection, and reliable visual perception under distribution shifts, with applications in automated driving and safety-critical systems.
\endbio

\end{document}

%% file: Images/Distribution_Transport.tex
%

\begin{tikzpicture}[
    every node/.style={font=\small},
    every node/.append style={transform shape}, 
    scale=0.78,
    axis/.style={->, thick},
    stem/.style={very thick},
    atom/.style={circle, fill, inner sep=1.6pt},
    transport/.style={->, thick, gray!75, >=Latex},
    dashedtransport/.style={->, thick, dashed, gray!70, >=Latex},
    brace/.style={decorate, decoration={brace, amplitude=5pt}},
]

\node[font=\bfseries] at (6.2,3.4) {(a) Transport of mass to extend the support};

\begin{scope}[shift={(0,0)}]
    \draw[axis] (0,0) -- (4.3,0) node[right] {$i$};
    \draw[axis] (0,0) -- (0,2.5) node[above] {$p(i)$};

    \node[blue] at (2.1,2.85) {Source distribution $p$};

    \foreach \x/\h/\lab in {0.6/1.0/{x_1},1.7/2.0/{x_2},2.7/0.6/{x_3},3.6/1.3/{x_4}}{
        \draw[stem, blue] (\x,0) -- (\x,\h);
        \node[atom, blue] at (\x,\h) {};
        \node[below] at (\x,0) {$\lab$};
    }

    \node[left] at (0,2.0) {$1$};
    \node[left] at (0,0) {$0$};

    \draw[brace, blue] (0.45,-0.55) -- (3.75,-0.55)
        node[midway, below=6pt] {$\mathrm{supp}(p)=\{x_1,x_2,x_3,x_4\}$};
\end{scope}

\begin{scope}[shift={(8,0)}]
    \draw[axis] (0,0) -- (5.2,0) node[right] {$j$};
    \draw[axis] (0,0) -- (0,2.5) node[above] {$q(j)$};

    \node[red] at (2.7,2.85) {Target distribution $q$};

    \foreach \x/\h/\lab in {0.4/0.5/{y_1},1.2/1.3/{y_2},2.1/0.9/{y_3},3.0/2.0/{y_4},3.8/0.2/{y_5},4.6/1.0/{y_6}}{
        \draw[stem, red] (\x,0) -- (\x,\h);
        \node[atom, red] at (\x,\h) {};
        \node[below] at (\x,0) {$\lab$};
    }

    \node[left] at (0,2.0) {$1$};
    \node[left] at (0,0) {$0$};

    \draw[brace, blue] (0.25,-0.55) -- (2.25,-0.55)
        node[midway, below=6pt, align=center] {overlap with\\$\mathrm{supp}(p)$};

    \draw[brace, red] (2.75,-0.55) -- (4.85,-0.55)
        node[midway, below=6pt, align=center] {new atoms\\support extension};
\end{scope}

\node at (3.5,2.2) {Transport plan $\gamma$};

\draw[transport] (3.7,1.4) to[out=35,in=150] (8.4,0.5);
\draw[transport] (3.7,1.4) to[out=35,in=150] (9.2,1.3);
\draw[transport] (3.7,1.4) to[out=35,in=140] (11.0,2.0);
\draw[transport] (3.7,1.4) to[out=35,in=150] (11.8,0.2);

\draw[dashedtransport] (3.7,1.1) to[out=25,in=160] (10.1,0.9);

\node[draw, rounded corners, align=left, fill=white] at (6.2,0.75) {
    \tikz{\draw[transport] (0,0) -- (0.6,0);} mass transported\\
    \tikz{\draw[dashedtransport] (0,0) -- (0.6,0);} zero / no transport
};

\draw[gray!70] (-0.3,-1.55) -- (13.6,-1.55);

\node[font=\bfseries] at (6.2,-2.0) {(b) Mass redistribution when enlarging the support};

\begin{scope}[shift={(0,-5)}]
    \draw[axis] (0,0) -- (3.7,0) node[right] {$i$};
    \draw[axis] (0,0) -- (0,2.0) node[above] {$p(i)$};

    \node[blue] at (1.8,2.35) {Observed distribution $p$};

    \foreach \x/\h/\lab in {0.4/0.8/{x_1},1.3/1.6/{x_2},2.2/0.35/{x_3},3.0/0.85/{x_4}}{
        \draw[stem, blue] (\x,0) -- (\x,\h);
        \node[atom, blue] at (\x,\h) {};
        \node[below] at (\x,0) {$\lab$};
    }

    \node[left] at (0,1.6) {$1$};
    \node[left] at (0,0) {$0$};

    \draw[brace, blue] (0.25,-0.55) -- (3.25,-0.55)
        node[midway, below=6pt] {$\Omega=\{x_1,x_2,x_3,x_4\}$};
\end{scope}

\begin{scope}[shift={(4.1,-5)}]
    \draw[axis] (0,0) -- (5.2,0) node[right] {$k$};

    \node at (2.65,2.35) {Transport / redistribution};

    \foreach \x/\h/\lab in {0.3/0.8/{x_1},1.1/1.0/{x_2},2.0/0.35/{x_3},2.8/0.75/{x_4}}{
        \draw[stem, blue] (\x,0) -- (\x,\h);
        \node[atom, blue] at (\x,\h) {};
        \node[below] at (\x,0) {$\lab$};
    }

    \foreach \x/\h/\lab in {3.6/0.55/{z_1},4.2/0.55/{z_2},4.8/0.55/{z_3}}{
        \draw[stem, purple] (\x,0) -- (\x,\h);
        \node[atom, purple] at (\x,\h) {};
        \node[below] at (\x,0) {$\lab$};
    }

    \foreach \x in {0.3,1.1,2.8}{
        \draw[blue, dashed] (\x-0.09,0.8) rectangle (\x+0.09,1.55);
    }

    \foreach \x in {3.6,4.2,4.8}{
        \draw[purple, dashed] (\x-0.09,0) rectangle (\x+0.09,0.9);
    }

    \draw[transport] (1.1,1.55) to[out=50,in=120] (3.6,0.65);
    \draw[transport] (1.1,1.55) to[out=45,in=110] (4.2,0.65);
    \draw[transport] (2.8,1.2) to[out=40,in=100] (4.8,0.65);

    \draw[brace, blue] (0.05,-0.55) -- (3.05,-0.55)
        node[midway, below=6pt] {original support $\Omega$};

    \draw[brace, purple] (3.35,-0.55) -- (5.05,-0.55)
        node[midway, below=6pt, align=center] {extended support\\$\Omega'=\Omega\cup\{z_1,z_2,z_3\}$};
\end{scope}

\begin{scope}[shift={(10,-5)}]
    \draw[axis] (0,0) -- (3.8,0) node[right] {$k$};
    \draw[axis] (0,0) -- (0,2.0) node[above] {$q(k)$};

    \node[red] at (2.0,2.35) {Extended distribution $q$};

    \foreach \x/\h/\lab in {0.3/0.75/{x_1},0.9/1.6/{x_2},1.5/0.45/{x_3},2.1/0.8/{x_4}}{
        \draw[stem, red] (\x,0) -- (\x,\h);
        \node[atom, red] at (\x,\h) {};
        \node[below] at (\x,0) {$\lab$};
    }

    \foreach \x/\h/\lab in {2.7/0.8/{z_1},3.2/0.3/{z_2},3.6/0.7/{z_3}}{
        \draw[stem, purple] (\x,0) -- (\x,\h);
        \node[atom, purple] at (\x,\h) {};
        \node[below] at (\x,0) {$\lab$};
    }

    \node[left] at (0,1.6) {$1$};
    \node[left] at (0,0) {$0$};

    \draw[brace, blue] (0.15,-0.55) -- (2.25,-0.55)
        node[midway, below=6pt] {original support $\Omega$};

    \draw[brace, purple] (2.55,-0.55) -- (3.75,-0.55)
        node[midway, below=6pt, align=center] {new atoms\\added to support};
\end{scope}

\node[
    draw,
    rounded corners,
    align=center,
    text width=8.5cm,
    fill=white
] at (6.4,-7.75) {
    Probability mass is transported from the observed support to new atoms,\\
    thereby \textcolor{purple}{extending the support} and
    \textcolor{red}{redistributing} mass.
};

\end{tikzpicture}

%% file: Images/simplex_mse.tex



\tikzset{
  corner/.style   = {circle, fill=black, inner sep=0.9pt},
  sample/.style   = {circle, fill=blue!65, inner sep=0.8pt},
  pull/.style     = {->, line width=0.45pt, opacity=0.7, shorten >=1.5pt},
  tri/.style      = {line width=0.4pt},
  axislabel/.style= {font=\scriptsize}
}

\newcommand{\DrawSmallTriangle}{
  \coordinate (A) at (0,0);
  \coordinate (B) at (3,0);
  \coordinate (C) at (1.5,2.598); 
  \draw[tri] (A)--(B)--(C)--cycle;
  \node[corner,label=below left:{\scriptsize class 1}] at (A) {};
  \node[corner,label=below right:{\scriptsize class 2}] at (B) {};
  \node[corner,label=above:{\scriptsize class 3}] at (C) {};
}

\begin{tikzpicture}[scale=0.8]
  \DrawSmallTriangle

  \foreach \x/\y in {0.5/0.2,0.7/0.25,2.3/0.25,2.5/0.2,1.45/2.2,1.55/2.2}
    {\node[sample] at (\x,\y) {}; }

  \draw[pull,blue!65] (1.5,1.0) -- (0.7,0.25);
  \draw[pull,blue!65] (1.6,1.1) -- (2.3,0.25);
  \draw[pull,blue!65] (1.5,1.2) -- (1.5,2.2);

\end{tikzpicture}

%% file: Images/simplex_wasserstein.tex

\tikzset{
  corner/.style   = {circle, fill=black, inner sep=0.9pt},
  sample/.style   = {circle, fill=orange, inner sep=0.8pt},
  pull/.style     = {->, line width=0.45pt, opacity=0.7, shorten >=1.5pt},
  tri/.style      = {line width=0.4pt},
  axislabel/.style= {font=\scriptsize}
}

\newcommand{\DrawSmallTriangle}{
  \coordinate (A) at (0,0);
  \coordinate (B) at (3,0);
  \coordinate (C) at (1.5,2.598); 
  \draw[tri] (A)--(B)--(C)--cycle;
  \node[corner,label=below left:{\scriptsize class 1}] at (A) {};
  \node[corner,label=below right:{\scriptsize class 2}] at (B) {};
  \node[corner,label=above:{\scriptsize class 3}] at (C) {};
}

\begin{tikzpicture}[scale=0.8]
  \DrawSmallTriangle

  \foreach \x/\y in {1.3/1.0,1.7/1.0,1.4/1.2,1.6/1.3,1.5/1.1,1.2/1.3,1.8/1.4}
    {\node[sample] at (\x,\y) {}; }

  \draw[pull,orange] (1.0,1.1) -- (1.4,1.1);
  \draw[pull,orange] (2.0,1.1) -- (1.6,1.2);
  \draw[pull,orange] (1.5,0.8) -- (1.5,1.1);

\end{tikzpicture}

%% file: biblio.bib
@InProceedings{Chen2018_deeplab,
author="Chen, Liang-Chieh
and Zhu, Yukun
and Papandreou, George
and Schroff, Florian
and Adam, Hartwig",
editor="Ferrari, Vittorio
and Hebert, Martial
and Sminchisescu, Cristian
and Weiss, Yair",
title="Encoder-Decoder with Atrous Separable Convolution for Semantic Image Segmentation",
booktitle="Computer Vision -- ECCV 2018",
year="2018",
publisher="Springer International Publishing",
address="Cham",
pages="833--851",
isbn="978-3-030-01234-2"
}

@inproceedings{Chan2021_smiyc,
 author = {Chan, Robin and Lis, Krzysztof and Uhlemeyer, Svenja and Blum, Hermann and Honari, Sina and Siegwart, Roland and Fua, Pascal and Salzmann, Mathieu and Rottmann, Matthias},
 booktitle = {Proceedings of the Neural Information Processing Systems Track on Datasets and Benchmarks},
 editor = {J. Vanschoren and S. Yeung},
 pages = {},
 title = {SegmentMeIfYouCan: A Benchmark for Anomaly Segmentation},
 url = {https://datasets-benchmarks-proceedings.neurips.cc/paper_files/paper/2021/file/d67d8ab4f4c10bf22aa353e27879133c-Paper-round2.pdf},
 volume = {1},
 year = {2021}
}

@article{Mackay1992,
    author = {MacKay, David J. C.},
    title = {A Practical Bayesian Framework for Backpropagation Networks},
    journal = {Neural Computation},
    volume = {4},
    number = {3},
    pages = {448-472},
    year = {1992},
    month = {05},
    issn = {0899-7667},
    doi = {10.1162/neco.1992.4.3.448},
}

@inproceedings{Lakshminarayanan2017,
author = {Lakshminarayanan, Balaji and Pritzel, Alexander and Blundell, Charles},
title = {Simple and scalable predictive uncertainty estimation using deep ensembles},
year = {2017},
isbn = {9781510860964},
publisher = {Curran Associates Inc.},
address = {Red Hook, NY, USA},
booktitle = {Proceedings of the 31st International Conference on Neural Information Processing Systems},
pages = {6405–6416},
numpages = {12},
series = {NIPS'17}
}

@misc{Mukhoti2020,
      title={Evaluating Bayesian Deep Learning Methods for Semantic Segmentation}, 
      author={Jishnu Mukhoti and Yarin Gal},
      year={2019},
      eprint={1811.12709},
      archivePrefix={arXiv},
      primaryClass={cs.CV},
      url={https://arxiv.org/abs/1811.12709}, 
}

@misc{Hendrycks2016,
  author = {Dan Hendrycks and
            Kevin Gimpel},
  title  = {A Baseline for Detecting Misclassified and Out-of-Distribution Examples
            in Neural Networks},
  year={2018},
      eprint={1610.02136},
      archivePrefix={arXiv},
      primaryClass={cs.NE},
doi={10.48550/arXiv.1610.02136}
}

@inproceedings{Maag2024_grads,
   title={Pixel-Wise Gradient Uncertainty for Convolutional Neural Networks Applied to Out-of-Distribution Segmentation},
   DOI={10.5220/0012353300003660},
   booktitle={Proceedings of the 19th International Joint Conference on Computer Vision, Imaging and Computer Graphics Theory and Applications},
   publisher={SCITEPRESS - Science and Technology Publications},
   author={Maag, Kira and Riedlinger, Tobias},
   year={2024},
   pages={112–122} 
}

@inproceedings{Sensoy2018_edl,
 author = {Sensoy, Murat and Kaplan, Lance and Kandemir, Melih},
 booktitle = {Advances in Neural Information Processing Systems},
 editor = {S. Bengio and H. Wallach and H. Larochelle and K. Grauman and N. Cesa-Bianchi and R. Garnett},
 pages = {},
 publisher = {Curran Associates, Inc.},
 title = {Evidential Deep Learning to Quantify Classification Uncertainty},
 url = {https://proceedings.neurips.cc/paper_files/paper/2018/file/a981f2b708044d6fb4a71a1463242520-Paper.pdf},
 volume = {31},
 year = {2018}
}

@INPROCEEDINGS{Vojir2023,
  author={Vojíř, Tomáš and Matas, Jiří},
  booktitle={IEEE/CVF Winter Conference on Applications of Computer Vision (WACV)}, 
  title={Image-Consistent Detection of Road Anomalies as Unpredictable Patches}, 
  year={2023},
  volume={},
  number={},
  pages={5480-5489},
doi={10.1109/WACV56688.2023.00545}
}

@inproceedings{Rai2023,
author = {Rai, Shyam and Cermelli, Fabio and Fontanel, Dario and Masone, Carlo and Caputo, Barbara},
year = {2023},
title = {Unmasking Anomalies in Road-Scene Segmentation},
booktitle    = {{IEEE/CVF} International Conference on Computer Vision, {ICCV}},
  pages        = {4014--4023},
  publisher    = {{IEEE}},
  doi          = {10.1109/ICCV51070.2023.00373},
}

@inproceedings{Delic2024,
author = {Anja Deli\'{c} and Matej Grcic and Sini\v{s}a \v{S}egvi\'{c}},
title     = {Outlier detection by ensembling uncertainty with negative objectness},
booktitle = {35th British Machine Vision Conference 2024, {BMVC} 2024, Glasgow, UK, November 25-28, 2024},
publisher = {BMVA},
year      = {2024},
url       = {https://papers.bmvc2024.org/0779.pdf}
}

@INPROCEEDINGS{Sodano2024,
author = { Sodano, Matteo and Magistri, Federico and Nunes, Lucas and Behley, Jens and Stachniss, Cyrill },
booktitle = { 2024 IEEE/CVF Conference on Computer Vision and Pattern Recognition (CVPR) },
title = { Open-World Semantic Segmentation Including Class Similarity },
year = {2024},
pages = {3184-3194},
doi = {10.1109/CVPR52733.2024.00307},
}

@conference{Marschall2025,
author={Samuel Marschall and Kira Maag},
title={Multi-Scale Foreground-Background Confidence for Out-of-Distribution Segmentation},
booktitle={International Joint Conference on Computer Vision, Imaging and Computer Graphics Theory and Applications},
year={2025},
pages={486-496},
doi={10.5220/0013241800003912},
}

@InProceedings{10.1007/978-3-031-26348-4_28,
author="Maag, Kira
and Chan, Robin
and Uhlemeyer, Svenja
and Kowol, Kamil
and Gottschalk, Hanno",
editor="Wang, Lei
and Gall, Juergen
and Chin, Tat-Jun
and Sato, Imari
and Chellappa, Rama",
title="Two Video Data Sets for Tracking and Retrieval of Out of Distribution Objects",
booktitle="Computer Vision -- ACCV 2022",
year="2023",
publisher="Springer Nature Switzerland",
address="Cham",
pages="476--494",
isbn="978-3-031-26348-4"
}

@article{Lee2018,
  author  = {Lee, Kimin and Lee, Kibok and Lee, Honglak and Shin, Jinwoo},
  year    = {2018},
  journal = {Neural Information Processing Systems (NeurIPS)},
  title   = {A Simple Unified Framework for Detecting Out-of-Distribution Samples and Adversarial Attacks}
}

@article{Liang2018,
  title   = {Enhancing The Reliability of Out-of-distribution Image Detection in Neural Networks},
  author  = {Shiyu Liang and Yixuan Li and Rayadurgam Srikant},
  journal = {International Conference on Learning (ICLR)},
  year    = {2018}
}

@inproceedings{Blum2019,
  author       = {Hermann Blum and
                  Paul{-}Edouard Sarlin and
                  Juan I. Nieto and
                  Roland Siegwart and
                  Cesar Cadena},
  title        = {Fishyscapes: {A} Benchmark for Safe Semantic Segmentation in Autonomous
                  Driving},
  booktitle    = {{IEEE/CVF} International Conference on Computer Vision Workshops,
                  {ICCV} Workshops},
  pages        = {2403--2412},
  publisher    = {{IEEE}},
  year         = {2019},
  doi          = {10.1109/ICCVW.2019.00294},
}

@INPROCEEDINGS {Cordts2016,
author = {M. Cordts and M. Omran and S. Ramos and T. Rehfeld and M. Enzweiler and R. Benenson and U. Franke and S. Roth and B. Schiele},
booktitle = {IEEE Conference on Computer Vision and Pattern Recognition (CVPR)},
title = {The Cityscapes Dataset for Semantic Urban Scene Understanding},
year = {2016},
volume = {},
issn = {1063-6919},
doi = {10.1109/CVPR.2016.350},
}

@inproceedings{Pinggera2016,
  author    = {Pinggera, Peter and Ramos, Sebastian and Gehrig, Stefan and Franke, Uwe and Rother, Carsten and Mester, Rudolf},
  booktitle = {IEEE/RSJ International Conference on Intelligent Robots and Systems (IROS)},
  title     = {Lost and Found: detecting small road hazards for self-driving vehicles},
  year      = {2016},
  volume    = {},
  number    = {},
pages={1099-1106},
  doi={10.1109/IROS.2016.7759186}
}

@inbook{Villani,
author = {Villani, Cédric},
year = {2008},
month = {01},
pages = {xxii+973},
title = {Optimal transport -- Old and new},
volume = {338},
doi = {10.1007/978-3-540-71050-9}
}

@article{peyre,
title = "Computational optimal transport",
author = "Gabriel Peyr{\'e} and Marco Cuturi",
note = "Publisher Copyright: {\textcopyright} 2019 Now Publishers Inc. All Rights Reserved.",
year = "2019",
month = jan,
day = "1",
doi = "10.1561/2200000073",
language = "English",
volume = "11",
pages = "1--257",
journal = "Foundations and Trends in Machine Learning",
issn = "1935-8237",
number = "5-6",
}

@inproceedings{brosch,
  author    = {Arnold Brosch and Abdelrahman Eldesokey and Michael Felsberg and Kira Maag},
  title     = {Out-of-Distribution Segmentation via Wasserstein-Based Evidential Uncertainty},
  booktitle = {European Symposium on Artificial Neural Networks, Computational Intelligence and Machine Learning (ESANN)},
  year      = {2025},
}

@inproceedings{kingma2015adam,
  author    = {Kingma, Diederik P. and Ba, Jimmy},
  title     = {Adam: A Method for Stochastic Optimization},
  booktitle = {International Conference on Learning Representations (ICLR)},
  year      = {2015},
}

@inproceedings{xie2021segformersimpleefficientdesign,
 author = {Xie, Enze and Wang, Wenhai and Yu, Zhiding and Anandkumar, Anima and Alvarez, Jose M. and Luo, Ping},
 booktitle = {Advances in Neural Information Processing Systems},
 editor = {M. Ranzato and A. Beygelzimer and Y. Dauphin and P.S. Liang and J. Wortman Vaughan},
 pages = {12077--12090},
 publisher = {Curran Associates, Inc.},
 title = {SegFormer: Simple and Efficient Design for Semantic Segmentation with Transformers},
 url = {https://proceedings.neurips.cc/paper_files/paper/2021/file/64f1f27bf1b4ec22924fd0acb550c235-Paper.pdf},
 volume = {34},
 year = {2021}
}

@INPROCEEDINGS {Wang_2023_CVPR,
author = { Wang, Dongdong and Gong, Boqing and Wang, Liqiang },
booktitle = { 2023 IEEE/CVF Conference on Computer Vision and Pattern Recognition (CVPR) },
title = {{ On Calibrating Semantic Segmentation Models: Analyses and An Algorithm }},
year = {2023},
volume = {},
ISSN = {},
pages = {23652-23662},
doi = {10.1109/CVPR52729.2023.02265},
url = {https://doi.ieeecomputersociety.org/10.1109/CVPR52729.2023.02265},
publisher = {IEEE Computer Society},
address = {Los Alamitos, CA, USA},
month =Jun}

@inproceedings{ArjovskyCB17,
  author       = {Mart{\'{\i}}n Arjovsky and
                  Soumith Chintala and
                  L{\'{e}}on Bottou},
  editor       = {Doina Precup and
                  Yee Whye Teh},
  title        = {Wasserstein Generative Adversarial Networks},
  booktitle    = {Proceedings of the 34th International Conference on Machine Learning,
                  {ICML} 2017, Sydney, NSW, Australia, 6-11 August 2017},
  series       = {Proceedings of Machine Learning Research},
  volume       = {70},
  pages        = {214--223},
  publisher    = {{PMLR}},
  year         = {2017},
  url          = {http://proceedings.mlr.press/v70/arjovsky17a.html},
  bibsource    = {dblp computer science bibliography, https://dblp.org}
}

@inproceedings{Frogner2015,
  title     = {Learning with a Wasserstein Loss},
  author    = {Frogner, Charlie and Zhang, Chiyuan and Mobahi, Hossein and Araya-Polo, Mauricio and Poggio, Tomaso},
  booktitle = {Advances in Neural Information Processing Systems (NeurIPS)},
  volume    = {28},
  year      = {2015},
  pages     = {2053--2061},
  url       = {https://papers.neurips.cc/paper/5679-learning-with-a-wasserstein-loss.pdf}
}

@inproceedings{Galesso2023,
  author        = {Silvio Galesso and Max Argus and Thomas Brox},
  title         = {Far Away in the Deep Space: Dense Nearest-Neighbor-Based Out-of-Distribution Detection},
  booktitle     = {Proceedings of the IEEE/CVF International Conference on Computer Vision (ICCV) Workshops},
  pages         = {4477--4487},
  year          = {2023},
}

@inproceedings{Vojir2024,
author = {Voj\'{\i}\v{r}, Tom\'{a}\v{s} and \v{S}ochman, Jan and Matas, Ji\v{r}\'{\i}},
title = {PixOOD: Pixel-Level Out-of-Distribution Detection},
year = {2024},
isbn = {978-3-031-73026-9},
publisher = {Springer-Verlag},
address = {Berlin, Heidelberg},
doi = {10.1007/978-3-031-73027-6_6},
booktitle = {Computer Vision – ECCV 2024: 18th European Conference, Milan, Italy, September 29–October 4, 2024, Proceedings, Part LX},
pages = {93–109},
numpages = {17},
}

@inproceedings{Ackermann2023,
author    = {Jan Ackermann and Christos Sakaridis and Fisher Yu},
title     = {Maskomaly: Zero-Shot Mask Anomaly Segmentation},
booktitle = {34th British Machine Vision Conference 2023, {BMVC} 2023, Aberdeen, UK, November 20-24, 2023},
publisher = {BMVA},
year      = {2023},
url       = {https://papers.bmvc2023.org/0329.pdf}
}

@inproceedings{Grcic2022,
  author    = {Grcic, Matej and Bevandić, Petra and Segvic, Sinisa},
editor       = {Shai Avidan and
                  Gabriel J. Brostow and
                  Moustapha Ciss{\'{e}} and
                  Giovanni Maria Farinella and
                  Tal Hassner},
  year      = {2022},
  volume       = {13685},
  pages        = {500--517},
  publisher    = {Springer},
  booktitle = {European Conference on Computer Vision (ECCV)},
  title     = {DenseHybrid: Hybrid Anomaly Detection for Dense Open-Set Recognition},
doi          = {10.1007/978-3-031-19806-9\_29},
}

@inproceedings{Tian2022,
  author    = {Yu Tian and
               Yuyuan Liu and
               Guansong Pang and
               Fengbei Liu and
               Yuanhong Chen and
               Gustavo Carneiro},
  title     = {Pixel-wise Energy-biased Abstention Learning for Anomaly Segmentation
               on Complex Urban Driving Scenes},
  year      = {2022},
  isbn = {978-3-031-19841-0},
    publisher = {Springer-Verlag},
    address = {Berlin, Heidelberg},
    doi = {10.1007/978-3-031-19842-7_15},
    booktitle = {Computer Vision – ECCV 2022: 17th European Conference, Tel Aviv, Israel, October 23–27, 2022, Proceedings, Part XXXIX},
    pages = {246–263},
    numpages = {18}
}

@inproceedings{Biase2021,
author       = {Giancarlo Di Biase and
                  Hermann Blum and
                  Roland Siegwart and
                  C{\'{e}}sar Cadena},
  title        = {Pixel-Wise Anomaly Detection in Complex Driving Scenes},
  booktitle    = {{IEEE} Conference on Computer Vision and Pattern Recognition, {CVPR}},
  pages        = {16918--16927},
  publisher    = {Computer Vision Foundation / {IEEE}},
  year         = {2021},
  doi          = {10.1109/CVPR46437.2021.01664},
}

@inproceedings{Gao2023,
 author = {Gao, Zhitong and Yan, Shipeng and He, Xuming},
 booktitle = {Advances in Neural Information Processing Systems},
 editor = {A. Oh and T. Naumann and A. Globerson and K. Saenko and M. Hardt and S. Levine},
 pages = {45150--45171},
 publisher = {Curran Associates, Inc.},
 title = {ATTA: Anomaly-aware Test-Time Adaptation for Out-of-Distribution Detection in Segmentation},
 url = {https://proceedings.neurips.cc/paper_files/paper/2023/file/8dcc306a2522c60a78f047ab8739e631-Paper-Conference.pdf},
 volume = {36},
 year = {2023}
}

@INPROCEEDINGS{Liu2023,
  author={Liu, Yuyuan and Ding, Choubo and Tian, Yu and Pang, Guansong and Belagiannis, Vasileios and Reid, Ian and Carneiro, Gustavo},
  booktitle={IEEE/CVF International Conference on Computer Vision (ICCV)}, 
  title={Residual Pattern Learning for Pixel-wise Out-of-Distribution Detection in Semantic Segmentation}, 
  year={2023},
  volume={},
  number={},
  pages={1151-1161},
  doi={10.1109/ICCV51070.2023.00112}
}

@inproceedings{Chan2021,
  author    = {Chan, Robin and Rottmann, Matthias and Gottschalk, Hanno},
  title     = {Entropy Maximization and Meta Classification for Out-of-Distribution Detection in Semantic Segmentation},
  booktitle = {IEEE/CVF International Conference on Computer Vision (ICCV)},
  year      = {2021},
publisher = {{IEEE}},
doi = {10.1109/iccv48922.2021.00508}
}

@article{Grcic2023,
  title={On Advantages of Mask-level Recognition for Outlier-aware Segmentation},
  author={Matej Grcic and Josip Šarić and Siniša Šegvić},
  journal={IEEE/CVF Conference on Computer Vision and Pattern Recognition Workshops (CVPRW)},
  year={2023},
  volume={},
  number={},
  pages={2937-2947},
  doi={10.1109/CVPRW59228.2023.00295}
}

@article{Nayal2024,
  author  = {Nazir Nayal and Youssef Shoeb and Fatma G{\"u}ney},
  title   = {A Likelihood Ratio-Based Approach to Segmenting Unknown Objects},
  journal = {International Journal of Computer Vision},
  volume  = {133},
  pages   = {6860--6872},
  year    = {2025},
  doi     = {10.1007/s11263-025-02509-0},
}

@inproceedings{Zhong2024,
  author        = {Liangyu Zhong and Joachim Sicking and Fabian Hueger and Hanno Gottschalk},
  title         = {{VL4AD}: Vision-Language Models Improve Pixel-Wise Anomaly Detection},
  booktitle     = {ECCV 2024 Workshop on Vision-Centric Autonomous Driving (VCAD)},
  year          = {2024},
}

@inproceedings{Holle2025,
  author        = {Marc H{\"o}lle and Walter Kellermann and Vasileios Belagiannis},
  title         = {Uncertainty-Aware Likelihood Ratio Estimation for Pixel-Wise Out-of-Distribution Detection},
  booktitle     = {Proceedings of the IEEE/CVF International Conference on Computer Vision (ICCV) Workshops},
  pages         = {772--782},
  year          = {2025},
}

@inproceedings{Nayal2023,
      title={RbA: Segmenting Unknown Regions Rejected by All}, 
      author={Nazir Nayal and Mısra Yavuz and João F. Henriques and Fatma Güney},
  booktitle    = {{IEEE/CVF} International Conference on Computer Vision, {ICCV}},
  pages        = {711--722},
  publisher    = {{IEEE}},
  year         = {2023},
  doi          = {10.1109/ICCV51070.2023.00072},
}

@InProceedings{Won2026,
      title={FlowCLAS: Enhancing Normalizing Flow Via Contrastive Learning For Anomaly Segmentation}, 
      author={Chang Won Lee and Selina Leveugle and Svetlana Stolpner and Chris Langley and Paul Grouchy and Jonathan Kelly and Steven L. Waslander},
      year={2026},
      booktitle={IEEE/CVF Winter Conference on Applications of Computer Vision (WACV)}, 
}

@inproceedings{Gudovskiy2023,
  author       = {Denis A. Gudovskiy and
                  Tomoyuki Okuno and
                  Yohei Nakata},
  editor       = {Robin J. Evans and
                  Ilya Shpitser},
  title        = {Concurrent Misclassification and Out-of-Distribution Detection for
                  Semantic Segmentation via Energy-Based Normalizing Flow},
  booktitle    = {Uncertainty in Artificial Intelligence, {UAI}},
  series       = {Proceedings of Machine Learning Research},
  volume       = {216},
  pages        = {745--755},
  publisher    = {{PMLR}},
  year         = {2023},
}

@article{Grcic2021,
  author = {Grcic, Matej and Bevandić, Petra and Kalafatić, Zoran and Segvic, Sinisa},
  title  = {Dense Out-of-Distribution Detection by Robust Learning on Synthetic Negative Data},
JOURNAL = {Sensors},
VOLUME = {24},
YEAR = {2024},
NUMBER = {4},
ARTICLE-NUMBER = {1248},
DOI = {10.3390/s24041248}
}

@inproceedings{Besnier2021,
  author        = {Victor Besnier and Andrei Bursuc and David Picard and Alexandre Briot},
  title         = {Triggering Failures: Out-of-Distribution Detection by Learning from Local Adversarial Attacks in Semantic Segmentation},
  booktitle     = {Proceedings of the IEEE/CVF International Conference on Computer Vision (ICCV)},
  pages         = {15681--15690},
  year          = {2021},
  doi           = {10.1109/ICCV48922.2021.01541},
}

@article{Lis2020,
  author = {Krzysztof Lis and
            Sina Honari and
            Pascal Fua and
            Mathieu Salzmann},
  title  = {Detecting Road Obstacles by Erasing Them},
journal      = {{IEEE} Transactions on Pattern Analysis and Machine Intelligence},
  volume       = {46},
  number       = {4},
  pages        = {2450--2460},
  year         = {2024},
  doi          = {10.1109/TPAMI.2023.3335152},
}

@inproceedings{Lis2019,
  author    = {Lis, Krzysztof and Nakka, Krishna and Fua, Pascal and Salzmann, Mathieu},
  year      = {2019},
  booktitle = {IEEE/CVF International Conference on Computer Vision (ICCV)},
  title     = {Detecting the Unexpected via Image Resynthesis},
pages        = {2152--2161},
  publisher    = {{IEEE}},
  doi          = {10.1109/ICCV.2019.00224},
}

@inproceedings{Vojir2021,
  author    = {Vojir, Tomas and \v{S}ipka, Tom\'a\v{s} and Aljundi, Rahaf and Chumerin, Nikolay and Reino, Daniel Olmeda and Matas, Jiri},
  title     = {Road Anomaly Detection by Partial Image Reconstruction With Segmentation Coupling},
  booktitle = {IEEE/CVF International Conference on Computer Vision (ICCV)},
  year      = {2021},
  volume={},
  number={},
  pages={15631-15640},
  doi={10.1109/ICCV48922.2021.01536}
}

@inproceedings{Zhang2024,
author = {Zhang, Hao and Li, Fang and Qi, Lu and Yang, Ming-Hsuan and Ahuja, Narendra},
title = {CSL: class-agnostic structure-constrained learning for segmentation including the unseen},
year = {2024},
isbn = {978-1-57735-887-9},
publisher = {AAAI Press},
doi = {10.1609/aaai.v38i7.28535},
booktitle = {Proceedings of the Thirty-Eighth AAAI Conference on Artificial Intelligence and Thirty-Sixth Conference on Innovative Applications of Artificial Intelligence and Fourteenth Symposium on Educational Advances in Artificial Intelligence},
articleno = {787},
numpages = {9},
series = {AAAI'24/IAAI'24/EAAI'24}
}

@InProceedings{Bevandic,
author="Bevandi{\'{c}}, Petra
and Kre{\v{s}}o, Ivan
and Or{\v{s}}i{\'{c}}, Marin
and {\v{S}}egvi{\'{c}}, Sini{\v{s}}a",
editor="Fink, Gernot A.
and Frintrop, Simone
and Jiang, Xiaoyi",
title="Simultaneous Semantic Segmentation and Outlier Detection in Presence of Domain Shift",
booktitle="Pattern Recognition",
year="2019",
publisher="Springer International Publishing",
address="Cham",
pages="33--47",
isbn="978-3-030-33676-9"
}

@article{Hein,
  author       = {Matthias Hein and
                  Maksym Andriushchenko and
                  Julian Bitterwolf},
  title        = {Why ReLU networks yield high-confidence predictions far away from
                  the training data and how to mitigate the problem},
  journal      = {CoRR},
  volume       = {abs/1812.05720},
  year         = {2018},
  url          = {http://arxiv.org/abs/1812.05720},
  eprinttype   = {arXiv},
  eprint       = {1812.05720},
  bibsource    = {dblp computer science bibliography, https://dblp.org}
}

@inproceedings{he2016residual,
  author = {He, Kaiming and Zhang, Xiangyu and Ren, Shaoqing and Sun, Jian},
  booktitle = {Proceedings of 2016 IEEE Conference on Computer Vision and Pattern Recognition},
  doi = {10.1109/CVPR.2016.90},
  issn = {1063-6919},
  location = {Las Vegas, NV, USA},
  month = jun,
  pages = {770--778},
  publisher = {IEEE},
  series = {CVPR '16},
  title = {{Deep Residual Learning for Image Recognition}},
  url = {http://ieeexplore.ieee.org/document/7780459},
  year = 2016
}

@misc{redmon2015unified,
  author = {Redmon, Joseph and Divvala, Santosh and Girshick, Ross and Farhadi, Ali},
  description = {[1506.02640] You Only Look Once: Unified, Real-Time Object Detection},
  note = {cite arxiv:1506.02640},
  title = {You Only Look Once: Unified, Real-Time Object Detection},
  url = {http://arxiv.org/abs/1506.02640},
  year = 2015
}

@InProceedings{10.1007/978-3-031-31438-4_3,
author="Holmquist, Karl and Klas{\'e}n, Lena and Felsberg, Michael",
title="Evidential Deep Learning for Class-Incremental Semantic
Segmentation",
booktitle="Scandinavian Conference on Image Analysis",
year="2023",
pages="32--48"
}

@INPROCEEDINGS {Xiong_10677853,
author = { Xiong, Ziliang and Jonnarth, Arvi and Eldesokey, Abdelrahman and Johnander, Joakim and Wandt, Bastian and Forssen, Per-Erik },
booktitle = { 2024 IEEE/CVF Conference on Computer Vision and Pattern Recognition Workshops (CVPRW) },
title = {{ Hinge-Wasserstein: Estimating Multimodal Aleatoric Uncertainty in Regression Tasks }},
year = {2024},
volume = {},
ISSN = {},
pages = {3471-3480},
doi = {10.1109/CVPRW63382.2024.00351},
url = {https://doi.ieeecomputersociety.org/10.1109/CVPRW63382.2024.00351},
publisher = {IEEE Computer Society},
address = {Los Alamitos, CA, USA},
month =Jun}
